\documentclass{article}

\usepackage{arxiv}

\usepackage[utf8]{inputenc} 
\usepackage[T1]{fontenc}    
\usepackage{hyperref}       
\usepackage{url}            
\usepackage{booktabs}       
\usepackage{amsfonts}       
\usepackage{nicefrac}       
\usepackage{microtype}      
\usepackage{lipsum}
\usepackage{graphicx}
\usepackage{multicol}
\usepackage{float}
\graphicspath{ {./images/} }

\title{Unlocking Legal Knowledge with Multi-Layered Embedding-Based
Retrieval}

\author{
 João Alberto de Oliveira Lima \\
  University of Brasília; Federal Senate of Brazil\\
  \texttt{joaolima@senado.leg.br} 
}

\begin{document}
\maketitle
\begin{abstract}
This work addresses the challenge of capturing the complexities of legal
knowledge by proposing a multi-layered embedding-based retrieval method
for legal and legislative texts. Creating embeddings not only for
individual articles but also for their components (paragraphs, clauses)
and structural groupings (books, titles, chapters, etc), we seek to
capture the subtleties of legal information through the use of dense
vectors of embeddings, representing it at varying levels of granularity.
Our method meets various information needs by allowing the Retrieval
Augmented Generation system to provide accurate responses, whether for
specific segments or entire sections, tailored to the user's query. We
explore the concepts of \emph{aboutness}, \emph{semantic chunking}, and
\emph{inherent hierarchy within legal texts}, arguing that this method
enhances the legal information retrieval. Despite the focus being on
Brazil's legislative methods and the Brazilian Constitution, which
follow a \emph{civil law} tradition, our findings should in principle be
applicable across different legal systems, including those adhering to
\emph{common law} traditions. Furthermore, the principles of the
proposed method extend beyond the legal domain, offering valuable
insights for organizing and retrieving information in any field
characterized by information encoded in hierarchical text.
\end{abstract}



\begin{multicols}{2}

\section{Introduction}
The increasing volume and complexity of legal corpora pose significant
challenges for legal professionals, including those in legislative
consultancy, where the efficient access and analysis of legal texts are
critical. Traditional keyword-based search methods often fall short in
capturing the nuances of legal language and the intricate relationships
within legal documents (\cite{Saravanan2009, Mimouni2014}).

Recent advancements in Generative Artificial Intelligence (GenAI) and
Retrieval Augmented Generation (RAG) systems offer promising avenues for
more efficient and accurate legal information retrieval. Embeddings,
which are dense and compact vector representations of text, effectively
capture the meanings of words, phrases, or documents (\cite{Yu2015,Chersoni2021}). Legislative documents, including bills and
legal normative statutes, inherently have a hierarchical structure. This
intrinsic hierarchy calls for an approach with variable granularity,
capable of representing both smaller segments and broader groupings of
legal texts through embeddings.

This paper proposes a multi-layered embedding-based retrieval method
that captures the semantic content of legal texts. By creating
embeddings for articles,\footnote{In Brazilian law, `article' is the
  fundamental unit within legislative texts, while in U.S. law,
  `Section' within codes and statutes are the comparable unit of
  division that detail specific provisions of the law. In this work we
  will use the term `article' as the basic unit of articulation, akin to
  a `section' in U.S. legislation (see Figure 3). We refrain from using
  the term `section' for the basic units since its literal translation
  in Portuguese (`seção') actually denotes a grouping of the basic units
  (`artigos').} their components, and their structural groupings, we aim
to provide a more nuanced and comprehensive representation of legal
knowledge. Our proposed approach enables RAG models to respond to user
queries with varying levels of detail, ranging from small portions to
comprehensive sections.

\section{Fundamental Concepts}

Before moving forward with an exploration of our method and its
applications, it's essential to define the \emph{core concepts} pivotal
to this study: \emph{embeddings}, \emph{aboutness}, \emph{retrieval
augmented generation}, and \emph{chunking}. These foundational elements
form the base of our methodology. Our goal is to ensure
clarity in our discourse for IT analysts and stakeholders within the
legal sector alike, who may not be well-versed in this specialized
terminology. By clarifying these key concepts, we aim to eliminate any
potential knowledge gaps.

\textbf{Embeddings:} At the heart of our approach lie embeddings, dense
vector representations of text that encapsulate semantic meanings within
documents. These vectors position words, sentences, or entire
documents\footnote{When dealing with texts that exceed the maximum token
  limit of the language model used for embedding, it becomes necessary
  to segment the text into semantic chunks. Each chunk is then processed
  to calculate its embedding. To represent the entire text, the
  arithmetic mean of these embedding vectors is computed, ensuring a
  cohesive representation despite the token limitation.} in a
multidimensional space, where the spatial proximity between points
signifies semantic similarity. This allows us to obtain a measure of the
semantic distance between texts (words, sentences or documents). Distant
vectors imply the texts that they stand for are dissimilar, while close
vectors imply they are similar.

In the legal domain, the intricate subtleties among various
terminologies are captured through embeddings, which can be visualized
in a two-dimensional or three-dimensional space via a process that
reduces the dimensionalities of the original embedding. Let's examine an
example using the fundamental concepts of Wesley Hohfeld's
framework.\footnote{For more information about Hohfeld\textquotesingle s
  framework, see \cite{DeOliveiraLima2021}.}

Figure 1 provides the source text for each embedding, alongside the
resulting vector representations. The textual descriptions define the
legal terms `Right', `Duty', `Power', and so on, outlining their
correlatives and opposites. These definitions were transformed into
dense numerical vectors, as shown in the last column, capturing the
semantic essence of each term. The vectors, which are the outcomes of
the embedding process, allow for the spatial mapping and analysis of
these concepts in a multidimensional vector space.
\end{multicols}

\textbf{Figure} 1. Hohfeld's Legal Positions Definitions and
Corresponding Embedding Vectors.

\includegraphics[width=6.0556in,height=2.60208in]{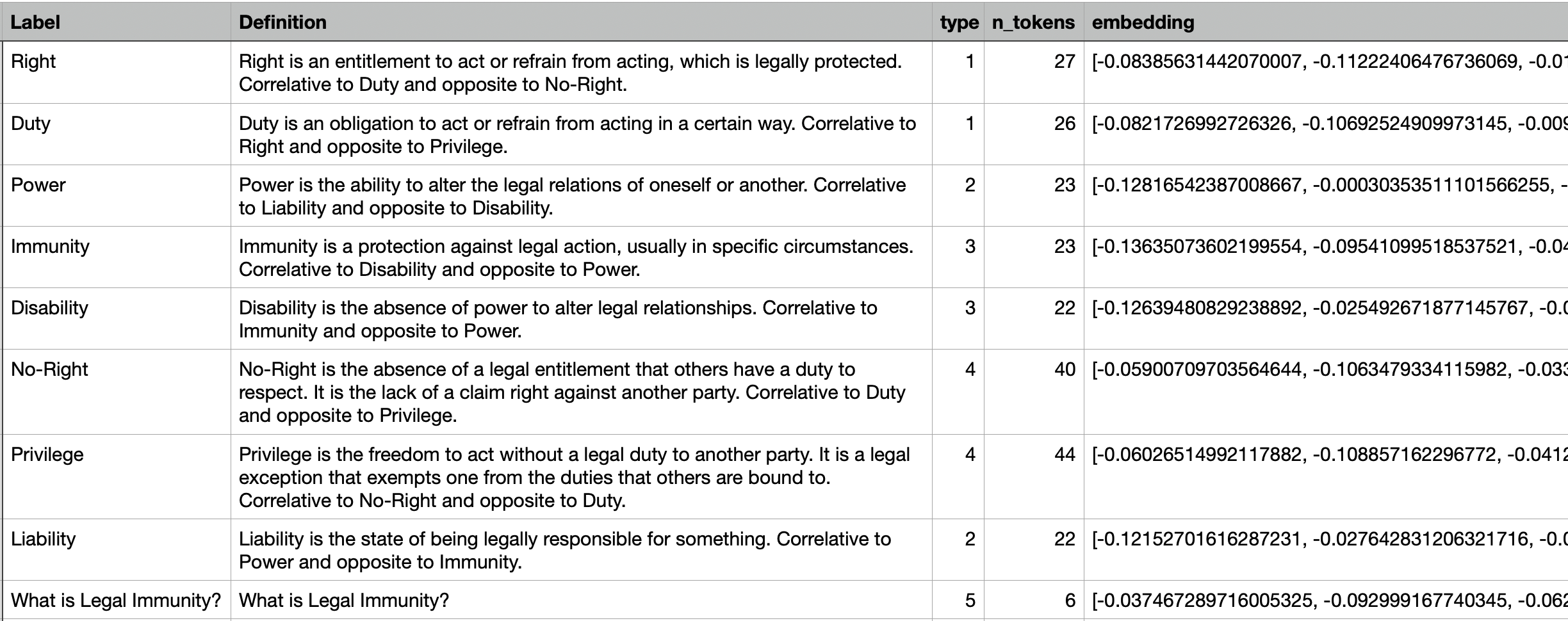}
\begin{multicols}{2} 
Figure 2 represents these embeddings in three dimensions and shows the
graph projection using the Plotly library\footnote{Plotly
  (\url{https://plotly.com/python/}) is an interactive, open-source
  graphing library for Python that enables users to create visually
  appealing, dynamic, and interactive visualizations. Plotly supports a
  range of charts and diagrams, including line charts, scatter plots,
  and complex three-dimensional models, which are essential for data
  analysis and presentation.}. It was necessary to reduce the original
3,072 dimensions to 3 dimensions using the PACMAP algorithm\footnote{Pacmap
  (https://github.com/YingfanWang/PaCMAP) is a dimensionality reduction
  algorithm that preserves the global data topology, ensuring that the
  relative distances in the high-dimensional space are meaningfully
  maintained in the reduced space, thus allowing for a more accurate
  representation of complex data structures.}. The first image shows the
projection of legal positions, while the second image details the cosine
similarity between these positions and between the positions and two
specific questions. In the second image, green outlines show the
semantic proximity between opposing and correlated legal positions.
Similarly, outlines in blue indicate the close relationship between the
two questions in the model and the legal positions they query. For
example, for the question `What is Legal Immunity?', the closest vector
is `Immunity'.

\end{multicols}
\pagebreak
\textbf{Figure 2.} 3D Map of Hohfeld's Legal Positions and Similarity
Matrix

\includegraphics[width=5.90556in,height=2.40208in]{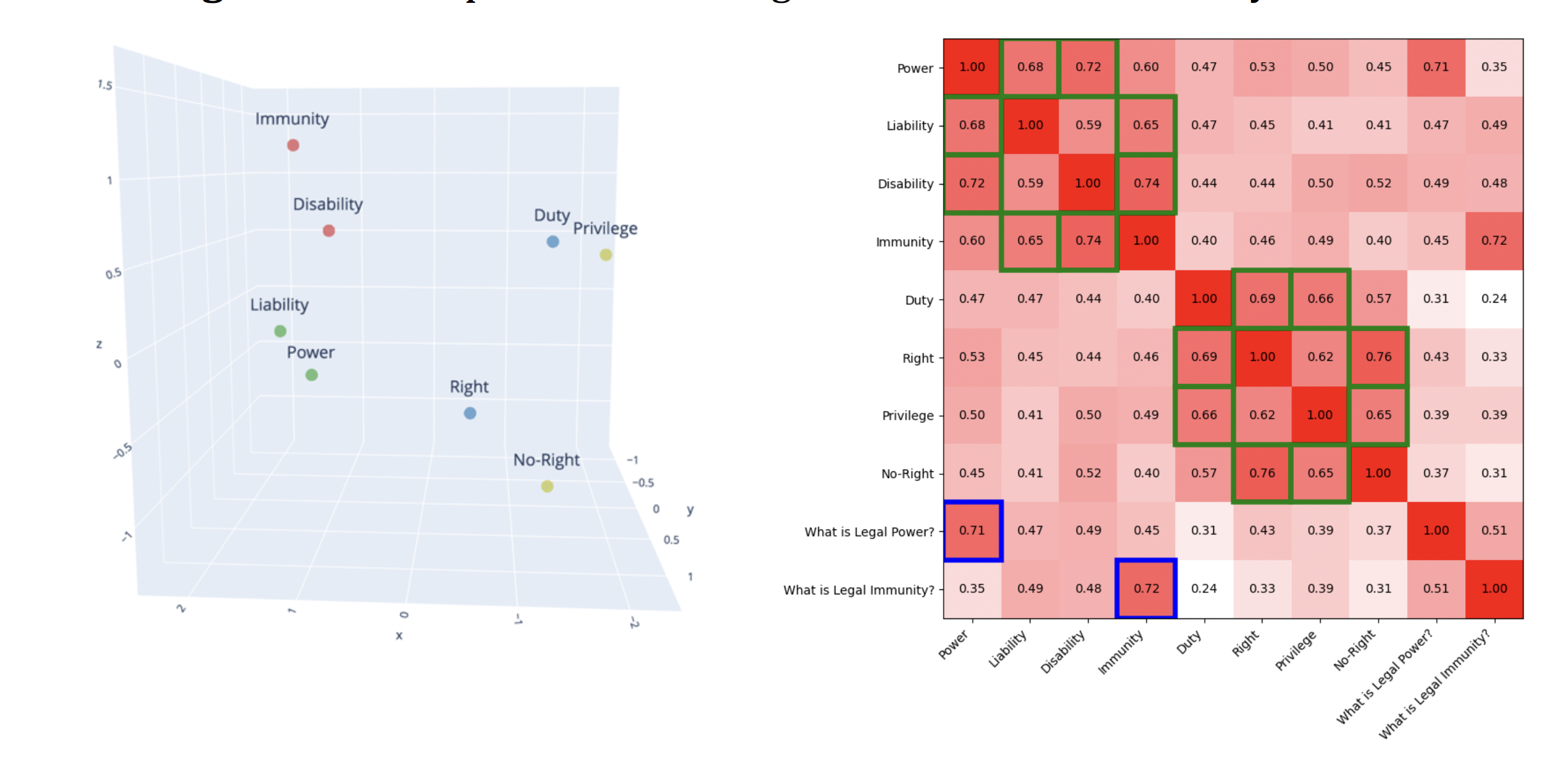}

\begin{multicols}{2} 

Beyond capturing the contextual meaning of text, embeddings also reflect
some syntactic features of documents. Texts that are semantically
identical but presented with slight formal variations---such as
differences in capitalization or line breaks---result in distinct yet
close vectors. This attention to semantic and syntactic features shows
how embeddings comprehensively capture a wide range of linguistic
nuances.

Moreover, it's important to challenge the conventional reliance on
keyword-based retrieval methods such as TF/IDF \cite{BenAyed2020,Huang2024}, which primarily focus on the frequency of
term occurrences. Unlike these traditional techniques, embeddings
provide a more precise approach by concentrating on the \emph{semantic
content} of texts. This allows discerning the underlying meanings and
themes, rather than merely counting words. This semantic focus makes
embeddings particularly effective in complex fields like law, where
exact terms may vary but the overarching legal concepts and
relationships remain consistent.

Before exploring further into the other concepts, one might wonder about
the intrinsic nature of an embedding. How does it capture and reflect
the essence of the text it represents? This question leads us to a
pivotal realization: embeddings can be seen as the digital ``aboutness''
of the content they encode.

\textbf{Aboutness:} The concept of aboutness refers to the main theme or
subject matter addressed by a text. It encompasses entities, concepts,
and relationships, and provides a comprehensive overview of its content.
Identifying a document's aboutness is necessary for pinpointing relevant
information in response to specific queries, thereby facilitating the
precise retrieval of legal information.

In examining the foundational concept of `aboutness', it is valuable to
reference Gilbert Ryle's early analysis of the term in his article
``About,'' from Analysis in 1933. Ryle differentiates between
interpretations of `aboutness', particularly emphasizing the `about
conversational' aspect. He articulates that an expression captures the
central theme of a conversation if a considerable number of sentences
within that conversation either directly use that expression, employ a
synonym, or make an indirect reference to the concept it denotes. This
occurrence should dominate without a competing expression or referent
being more prevalent across the dialogue. Ryle's viewpoint suggests that
`aboutness' serves as a marker for the primary concern within a
discourse, indicated through frequent references or mentions, rather
than through the mere presence of specific keywords \cite{Ryle1933}.

This consideration of Ryle's perspective aids in refining our approach
to embedding legal documents. Recognizing `aboutness' as more than a
function of explicit text, but rather as a reflection of the thematic
consistency across a document, informs our strategy. It acknowledges
that the core theme or subject matter of a discourse is not necessarily
tied to the explicit repetition of certain terms but is instead often
conveyed through varied linguistic mechanisms that signal ongoing
engagement with the theme.

\textbf{Retrieval Augmented Generation (RAG)} is a method that enhances
response formulation by leveraging data as a foundational source. This
technique capitalizes on both traditional access methods and the
semantic prowess of embeddings, thus ensuring that the generated
responses are not only relevant but also contextually rich. The
framework is a structured approach divided into three main phases:
Indexing, Retrieval, and Generation.

In the \emph{indexing phase}, an extensive index is created using a
variety of external sources. If employing traditional approaches,
several steps are necessary for textual indexing, including processes
like text normalization, which encompasses tokenization and stemming,
along with the removal of stopwords, to refine the data for enhanced
searchability. Alternatively, advanced approaches involve using
pretrained language models to produce semantic vectors of texts
(embeddings). In both approaches, it is essential to decide on the
indexing unit , whether it be the entire document, uniform parts of the
document, or other criteria.

The \emph{retrieval phase} utilizes the index to find information
relevant to user queries. In traditional methods, this process looks at
term frequency and the presence of specific terms to rank documents.
Common techniques such as TF/IDF or BM25 prioritize documents based on
how frequently the search terms appear. Conversely, when using
embeddings within the RAG framework, this phase includes selecting
indexed items based on their semantic proximity to user questions,
employing vector similarity techniques such as cosine similarity or
Euclidean distance. This advanced approach allows evaluating the
semantic nuances of queries, enhancing ranking by closely aligning
results with user intentions and the context of the query. The
mathematical representation of user queries through embeddings creates a
certain language independence in that the same query expressed by
similar words (synonyms or near-synonyms), or even in different
languages, can achieve similar results.

The \emph{generation phase} involves creating responses by combining the
retrieved information (concatenated chunks as context) with the original
user query. This process typically uses a large language model to ensure
the response is contextually relevant and reflects the contextual
retrieved data accurately.

To move from the broader framework of RAG to the specifics of the
\emph{indexing phase}, we must first focus on the technique of chunking,
which plays a key role in organizing and processing text for retrieval.

\textbf{Chunking:} This process involves dividing text into smaller,
manageable segments, such as sentences, paragraphs, or specific legal
clauses. Segmenting complex documents into parts allows for a granular
examination of each segment's content and its contribution to the
document's overarching meaning, thus facilitating the analysis and
processing of intricate legal texts.

Semantic chunking, as practiced in semchunk\footnote{\url{https://github.com/umarbutler/semchunk}
  .}, offers a method for dividing text into smaller, semantically
coherent segments, utilizing a recursive division technique that
initially splits text using contextually meaningful separators. This
process is repeated until all segments reach a specific size, ensuring
each chunk represents a unique thematic or conceptual element. However,
this approach overlooks the intrinsic hierarchical organization of
legislative texts, which are structured in a complex hierarchy where the
systematization of provisions is explicitly defined by the author. While
traditional semantic chunking effectively creates manageable segments
using only one layer, it overlooks the multitude of layers derived from
the hierarchical organization. In other words, it treats all segments on
the same level, disregarding the structured organization that defines
their legal significance and their interrelations.

A clear grasp of the aforementioned concepts is essential before
exploring the specific characteristics of articulated normative texts.
The ensuing section will elaborate on how these principles are applied
to represent the complex, hierarchical nature of legislative texts,
enabling more effective retrieval of legal knowledge.

\section{The Specificity of Legal and Legislative
Information}
\label{the-specificity-of-legal-and-legislative-information}

A defining characteristic of legislative texts is an inherent
hierarchical structure. It reflects the systematization of legal
dispositions, ranging from broad legal areas to specific articles,
clauses, and sub-clauses.

In structured normative texts such as constitutions, every detail has
substantial semantic importance. Unlike ordinary language, where
redundancy and informality are common, constitutions are carefully
crafted documents where each word, phrase, and clause is deliberately
chosen to convey specific legal meaning and effect. This precision
requires a granular approach to information retrieval. Analyzing and
representing legal knowledge at varying levels of detail -- from
individual clauses and sub-clauses to broader articles, chapters and
titles -- allows us to capture the full spectrum of meaning embedded
within these texts.

Figure 3 depicts the organizational structure of legislative documents
in Brazil, conforming to the guidelines of Complementary Law No.
95/1998. The highest level is the ``Legal Norm'' (Norma Jurídica), under
which the Main Text (Texto Principal) is the primary body of the
document. Optionally, one or more Annex (Anexo) may be included,
indicated by a dashed line, suggesting its non-mandatory nature. It is
important to note that the number of Annex can vary, ranging from zero
to several, indicating the flexibility for including them.

The foundational unit of this framework is the ``Article,'' which
consists of a mandatory ``Lead paragraph'' (``\emph{caput}'') and,
potentially, one or more ``Paragraphs'' (``Parágrafo''), indicated by
dashed lines to denote their optionality. These ``Paragraphs'' elucidate
details or exceptions pertaining to the ``Lead paragraph.'' If
enumeration is necessary, the lead paragraph or paragraphs can branch
into ``Sections'' (incisos), which may be further divided into ``Items''
(alíneas), and then into ``Subitems'' (item) for detailed specificity.

Simple statutes may consist of only articles, but complex statutes and
codes group articles into progressively larger units: ``Sections
{[}group of Articles{]}'' and ``Subsections'' are possible components of
``Chapters,'' which are compiled under ``Titles.'' These titles are then
organized into ``Books,'' and multiple books can be part of a larger
``Part,'' reflecting the elaborate nature of legal codes.

\end{multicols}
\textbf{Figure 3}. Internal hierarchy of a Brazilian statute, detailing
its constituent elements.

\includegraphics[width=5.90556in,height=3.16875in]{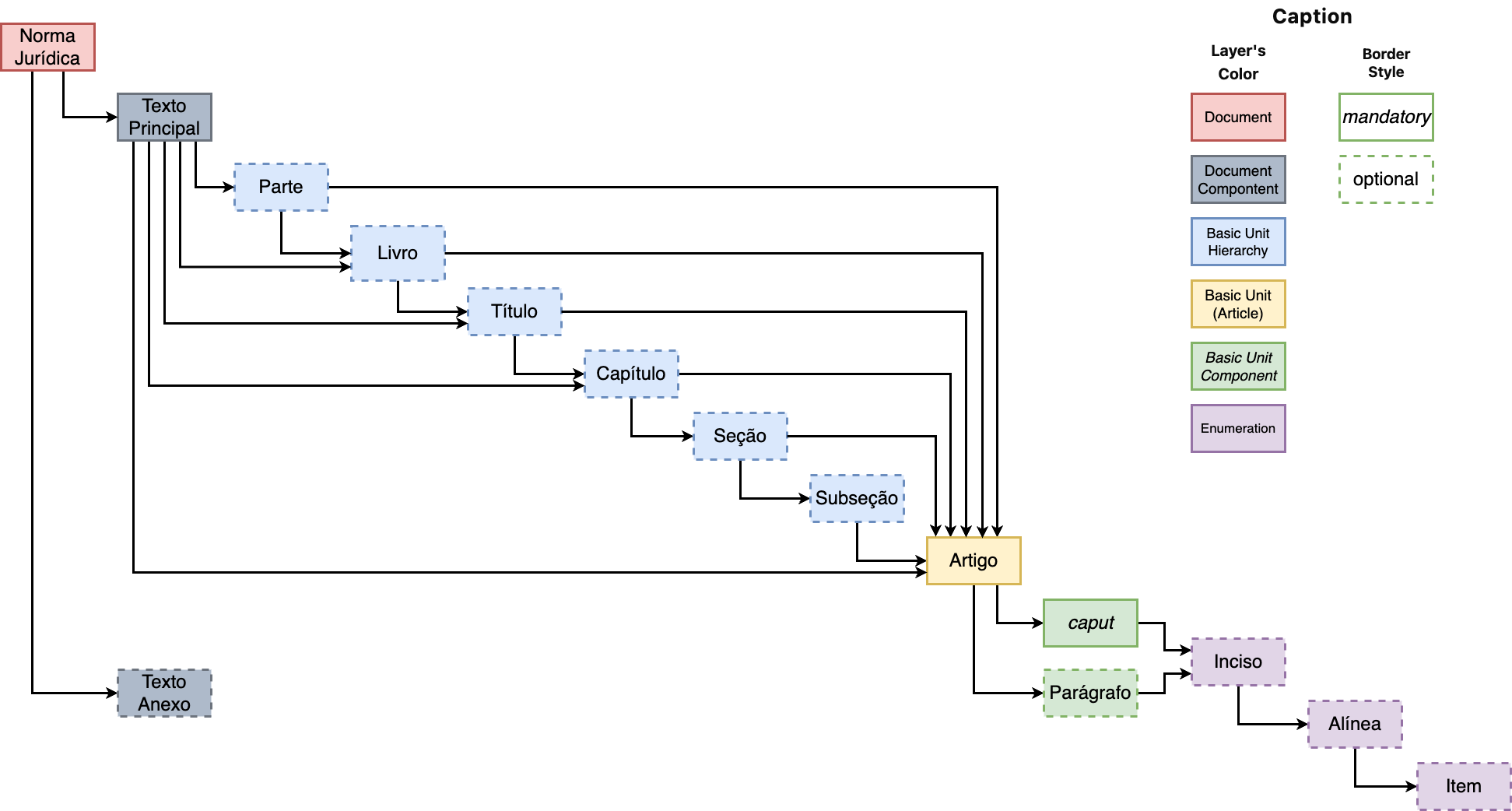}
\begin{multicols}{2}

Figure 4 provides an overview of the hierarchical structure of Title I
and Chapter I of Title II of the Brazilian Constitution. It visually
represents how the constitutional text is organized, starting from the
broadest divisions and narrowing down to specific provisions. Each
``Article'' serves as a foundational element, with the ``Lead
paragraph'' (caput) stating the principal point. ``Sole Paragraphs''
(parágrafo único) provide further explanations or stipulations when an
article has only one. For detailed enumeration within an article,
``Sections'' (incisos) break down the content into subsections, each
identified by Roman numerals. These sections can further subdivide into
``Items'' (alíneas), denoted by lowercase letters, and
\textquotesingle Subitems\textquotesingle{} for the most specific
details, though not depicted in this portion of the structure. This
systematic approach allows for a clear and organized reading of the
constitutional provisions.

\end{multicols}
\textbf{Figure 4.} Segment of the Hierarchical Structure of the
Brazilian Constitution

\includegraphics[width=4.92405in,height=3.95241in]{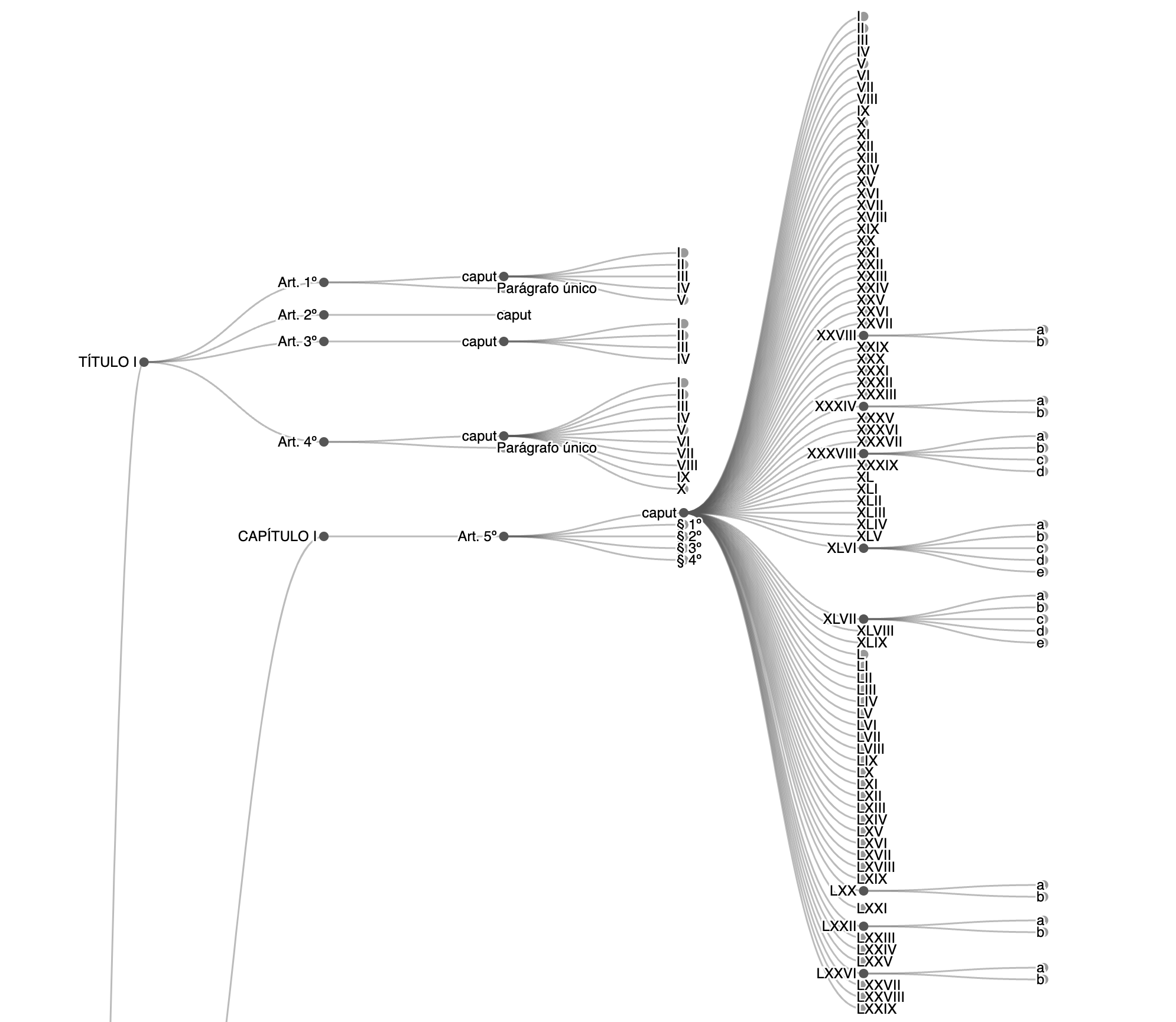}
\begin{multicols}{2}

Figure 5 is an extract from Title I of the Brazilian Constitution,
highlighting its hierarchical structure using color-coded annotations.
The Document Component layer indicates the beginning of the main text of
the Brazilian Constitution, while the Basic Unit Hierarchy layer
contains all the Titles, beginning with Title I. The Basic Unit layer
delineates Article 1, detailing the core values of the federation such
as sovereignty, citizenship, human dignity, social values of labor and
free enterprise, and political pluralism. The Basic Unit Component layer
focuses on the ``Lead paragraph'' (caput) of Article 1, which declares
Brazil as an indivisible federation and defines its democratic, legal
foundation. Lastly, the Enumeration layer specifically identifies
enumeration elements like Section III (inciso) of the lead paragraph of
Article 1, which underscores `human dignity' as a fundamental principle.
\end{multicols}
\textbf{Figure 5.} Layers of the beginning of the Brazilian Constitution

\includegraphics[width=6.5114in]{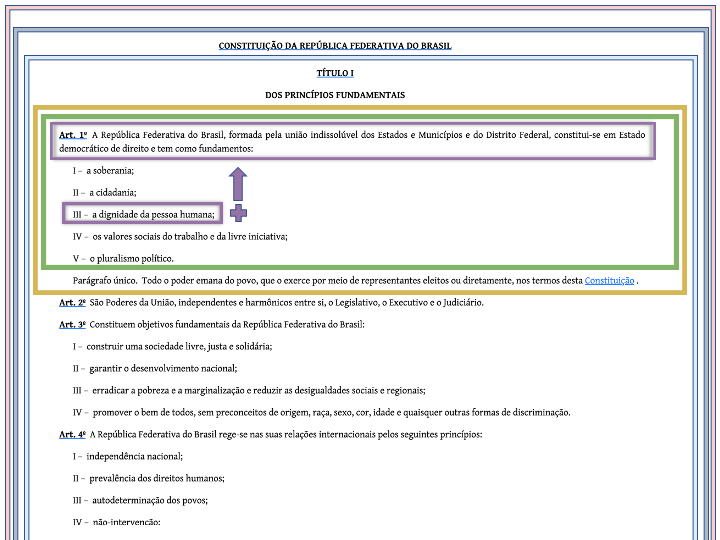}
\begin{multicols}{2}
Considering their fundamental role in systematizing legal norms and
principles, ``Articles'' are the most suitable unit for initial semantic
chunking within legislative texts. As illustrated in Figure 4, applying
this approach to Title I of the Brazilian Constitution results in four
distinct chunks, each representing an individual article. While this
provides a solid basis for analysis, our proposed methodology, detailed
in subsequent sections, seeks to further enhance this process.
Generating a broader set of chunks for embedding allows us to create a
more detailed and complete representation of legal knowledge, ultimately
leading to greater accuracy and effectiveness in legal information
retrieval.

Building on our understanding of the hierarchical nature of legislative
texts, we now present our proposal for a multi-layered embedding-based
retrieval method. This approach aims to capture the semantic content of
legal documents at various levels of granularity, from entire documents
to individual clauses. The following section will outline the specifics
of our chunking methodology in the indexing phase and the strategies for
filtering chunks in the retrieval phase.

\section{Proposal for Utilizing a Multi-Layered Embedding-Based Retrieval }

This section explores our proposed approach, offering recommendations
that cover two RAG phases: chunking methodology during indexing and
filtering chunks at retrieval.

\hypertarget{chunking-methodology-during-rags-indexing-phase}{%
\subsection{Chunking methodology during RAG's indexing
phase}\label{chunking-methodology-during-rags-indexing-phase}}

Our proposed solution defines the following layers for segmenting the
textual content of a legislative or legal document:

\begin{itemize}
\item
  Document Level: At the highest level, we generate one embedding for
  each document. This captures the overarching theme, purpose, and scope
  of the legal text. This comprehensive vector can then be used to
  perform automatic classification of the document, aligning it with
  relevant legal categories.
\item
  Document Component Level: This level concentrates on capturing the
  essence of document components, which can be systematically presented
  through articles (such as the main text) or depicted using various
  structures (like tables or unstructured text). Examples encompass the
  main text and additional texts components such as justifications (in
  bills) and schedules (in annexes), with each being assigned its own
  embedding to represent its distinct contribution to the overarching
  legal context.
\item
  Basic Unit Hierarchy Level: Embeddings are also generated for books,
  titles, chapters, and sections. This captures the broader themes and
  relationships between groups of articles.
\item
  Basic Unit Level: Each article, as a fundamental unit of legal text,
  receives its own embedding. This captures the specific legal issue
  addressed by the article and its core provisions.
\item
  Basic Unit Component Level: Further granularity is achieved by
  creating embeddings for the obligatory head or the paragraphs that
  compose each article. This allows a detailed understanding of the
  semantics in each legal provision.
\item
  Enumeration Level: Further granularity is achieved by creating
  embeddings for item that are part of an article head or a paragraph.
  This allows a detailed understanding of the specific details in each
  legal provision.
\end{itemize}

By incorporating embeddings at these different levels, we create a
multi-layered representation of legal knowledge, enabling RAG models to
respond to user queries with varying levels of detail.

To illustrate the granularity achieved by our proposed method, let us
consider the example presented in Figure 4. A standard approach using
articles as the embedding unit would segment this text into 4 chunks,
corresponding to the 4 articles. However, our hierarchical approach
yields a much richer representation. At the Document-Level, we have 1
chunk encompassing the entire text. Similarly, the Text Component-Level
also produces 1 chunk, as it encompasses the main body of the text. The
Article-Level retains the 4 chunks corresponding to the articles, but
the Article Component-Level breaks these down into another 25 chunks: 4
lead paragraphs (caputs), 2 sole paragraphs, and 19 sub-sections
(incisos). Finally, the Grouping-Level provides 1 chunk representing the
single title. In total, our method generates 32 chunks, thus providing a
significantly more detailed representation compared to the 4 chunks of
the traditional approach.

Figure 5 not only showcases the structure of Title I of the Brazilian
Constitution but also reveals an insightful detail regarding Article
Component-Level embedding within legal texts. Specifically, when it
comes to enumerative elements such as `Sections' (incisos), `Items'
(alíneas), and `Subitems' (itens), it is not sufficient to consider
their text in isolation. Since each element elaborates on a part of a
larger context, it is essential to embed their text within the context
of the superior elements up to the level of the `Lead paragraph' (caput)
of the article or the paragraph to which they belong. For instance,
instead of isolating the phrase \emph{``a dignidade da pessoa humana''}
(the dignity of the human person), the embedded text should read as
follows: \emph{``{[}A República Federativa do Brasil, formada pela união
indissolúvel dos Estados e Municípios e do Distrito Federal,
constitui-se em Estado democrático de direito e tem como fundamentos:{]}
dignidade da pessoa humana''}\footnote{``{[}The Federative Republic of
  Brazil, formed by the indissoluble union of the states and
  municipalities and of the Federal District, is a legal democratic
  state and is founded on:{]} the dignity of the human person;''}. This
method of embedding ensures that the specific provisions are always
understood within the full scope of their intended legal framework.

\hypertarget{filter-strategy-during-rags-retrieval-phase}{%
\subsection{Filter strategy during RAG's Retrieval
phase}\label{filter-strategy-during-rags-retrieval-phase}}

The strategy for filtering chunks in the retrieval system utilizes
cosine similarity alongside additional parameters to ensure precise and
contextually relevant responses. After computing the cosine similarities
between the query embedding and each chunk\textquotesingle s embedding
within the dataset, the results are sorted based on similarity scores,
prioritizing the highest values.

The selection process for the chunks includes several critical steps.
Firstly, a baseline token count is set, which in this example is 2,500
tokens. This limit helps manage the total volume of text being
considered, preventing overly lengthy or unmanageable responses.
Additionally, a baseline percentage for similarity deviation, set here
at 25\%, is used to compare each chunk's similarity against the highest
found similarity. This comparison helps to filter out chunks that are
significantly less similar to the query, maintaining a focus on those
most closely related in semantic content.

Chunks are also filtered based on their textual boundaries to avoid
overlap in the content presented. If a chunk's textual range is already
covered by another chunk with a higher similarity score, it is not
included again, ensuring diversity and relevance in the returned
results. This helps in maintaining a concise and focused response area.
For instance, if ``Art. 5, § 1'' is retrieved, there is no need to
retrieve ``Art. 5, § 1, I'' because the latter is encompassed by the
former.

The final selection of chunks also considers the total number of tokens
accumulated thus far against the baseline. After at least seven segments
have been chosen, the process of selecting additional chunks continues
until the total number of tokens exceeds the baseline threshold or the
similarity falls below 25\% of the highest similarity found, whichever
comes first. This method balances detail and breadth, providing a
comprehensive yet focused array of text for response generation and
optimizing both the quality and efficiency of the information retrieval
process.

To validate the effectiveness of our multi-layered embedding-based
retrieval method, we conducted a comparative analysis against the
traditional flat chunking approach. This analysis involved visualizing
the embeddings and testing the retrieval results using a series of
queries about the Brazilian Constitution. The following section presents
the results of this comparative study, highlighting the advantages and
potential limitations of our proposed method.

\hypertarget{comparative-of-flat-vs-multi-layered-chunking-approaches}{%
\section{Comparative of Flat vs Multi-Layered Chunking
Approaches}\label{comparative-of-flat-vs-multi-layered-chunking-approaches}}

This section will present a two-dimensional projection visualization of
embeddings and demonstrate a test implementation of RAG for the
Brazilian Constitution. We will compare flat (divided by articles) and
multi-layered chunking approaches. In the second part, we will test the
retrieval results of both approaches using six queries about provisions
of the Brazilian Constitution.

In this experiment, we employed the text-embedding-3-large
model\footnote{\url{https://platform.openai.com/docs/guides/embeddings}
  .}, configured to generate 256-dimensional vectors. To present the
data in a comprehensible manner, we reduced these high-dimensional
embeddings to a two-dimensional plane using the dimensionality reduction
technique (PACMAP). The visualizations were then created using Plotly,
the interactive graphing library, to facilitate a clearer understanding
of the data structures and their semantic correlations.

Figure 6 showcases a visualization of article embeddings (traditional
method of uniform fragments) of the provisions in Brazil's Federal
Constitution, displayed in a two-dimensional space for easier
comprehension. Each point represents an article plotted according to its
embedding in a simplified two-dimensional space. This offers a broad
overview of how each article is positioned relative to others,
potentially reflecting their semantic proximity and thematic linkages.
\end{multicols}
\textbf{Figure 6}. Article Embedding of the Brazilian Constitution
(highlighting Article 2)

\includegraphics[width=6.30759in]{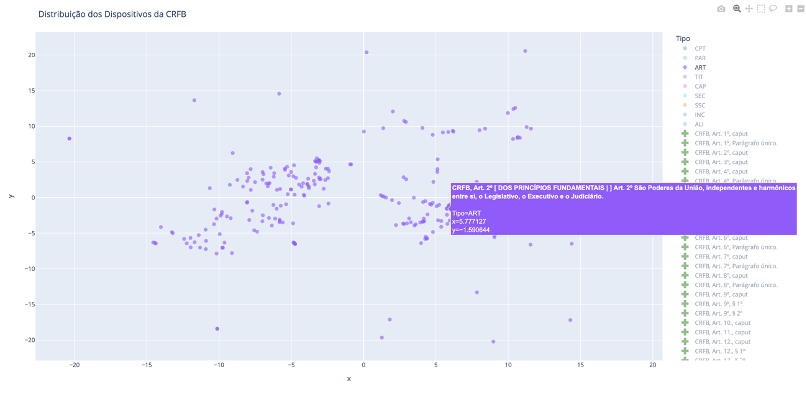}
\begin{multicols}{2}
Figure 7 exemplifies our hierarchical embedding approach, using the same
corpus of the Brazilian Constitution text. The visualization in Figure 7
illustrates the hierarchical embedding method\textquotesingle s capacity
to generate a multitude of proxies---2954 in total---that more precisely
correspond to the semantic content of the legal text, unlike the
traditional method which produces only 276. The visualization becomes
significantly more intricate. It includes embeddings from various levels
of granularity, as advocated in this paper. This comprehensive approach
incorporates not only the articles but also their components such as
paragraphs, clauses, and sub-clauses, as well as their groupings like
titles and chapters.
\end{multicols}
\textbf{Figure} 7. The Hierarchical Embedding of the Brazilian
Constitution (highlighting 4\textsuperscript{th} Paragraph of Article
14)

\includegraphics[width=6.30556in]{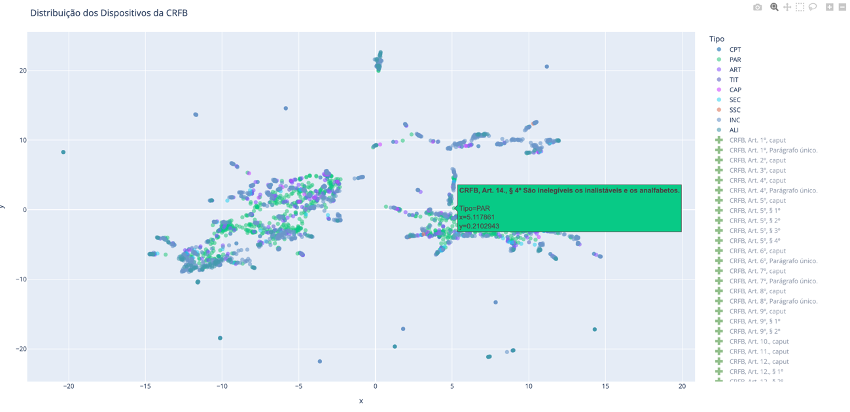}
\begin{multicols}{2}

Figure 8 takes a close look at the Article 5 region, such as its
numerous clauses and sub-clauses. Given that Article 5 contains over 70
such components, this figure demonstrates the increased density of
points surrounding the Article 5 embedding. It illustrates the detailed
semantic landscape that emerges when each subdivision is given a
distinct representation, underscoring the sophistication of the proposed
multi-layered embedding approach. In Figure 6, note that Article 5 is
represented by just a single point, whereas in the hierarchical view, it
resembles a cluster.
\end{multicols}
\textbf{Figure} 8. The Hierarchical Embedding of the Brazilian
Constitution Article 5 region

\includegraphics[width=6.30556in]{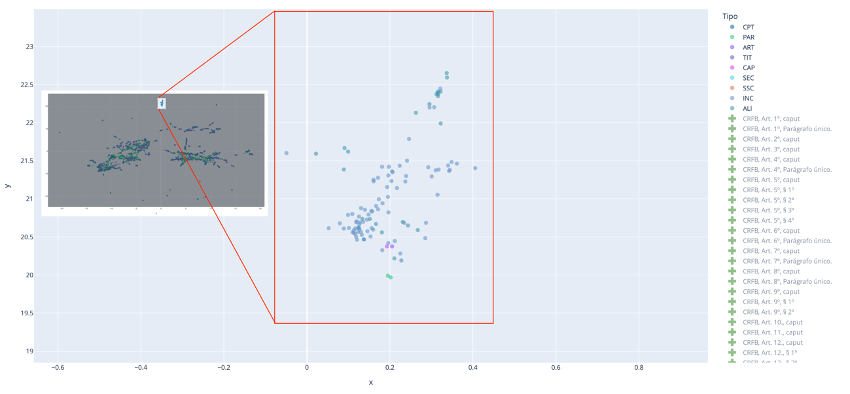}
\begin{multicols}{2}

When comparing Figures 6 and 8, it becomes evident that the latter, in
addition to possessing the vector representing Article 5 in its entirety
(purple), also contains dozens of other vectors representing more
specific aspects of this article that deals with fundamental rights.
This allows the RAG strategy to offer more relevant excerpts in a more
concise and consequently, more economical manner, if the question should
be answered based in fine grained provisions.

\section{Comparative of Flat vs Multi-Layered Embedding-Based Retrieval}

Table 1 presents the questions formulated for a simple comparative test
of response retrieval and generation using Multi-Layer (ML) and Flat
embedding approaches. In crafting these questions, we aimed to encompass
both specific and comprehensive inquiries. Appendix A lists all the
devised questions along with descriptive information about the selected
chunks and an evaluation of each chunk\textquotesingle s relevance to
the response, graded from \textquotesingle essential\textquotesingle{}
through \textquotesingle complementary\textquotesingle{} to
\textquotesingle unnecessary\textquotesingle. Additionally, we have
included the corresponding responses from the generation phase,
accompanied by a brief commentary on each response. In this comparative
test, the "gpt-4-turbo-preview" model was used during the generation
stage, limiting the output to 1000 tokens and setting the temperature at
0.3.
\end{multicols}
\begin{table}
\caption{Case Study Questions and Basic Statistics}
\begin{tabular}{p{0.5334\textwidth} r r r r}
\toprule
\textbf{Question} & \multicolumn{2}{c}{\textbf{Total Retrieved Chunks}} & \multicolumn{2}{c}{\textbf{Total Chunk Tokens}} \\
\cmidrule(lr){2-3} \cmidrule(lr){4-5}
& ML & Flat & ML & Flat \\
\midrule
Q1. What are the foundations of the republic? & 19 & 7 & 2,502 & 3,110 \\
Q2. Talk about the social function of property. & 12 & 11 & 2,586 & 2,625 \\
Q3. What are the attributes of the vote? & 7 & 7 & 2,966 & 4,549 \\
Q4. How is tax revenue distributed? & 8 & 7 & 2,796 & 3,595 \\
Q5. What are the rights of children and teenagers? & 7 & 7 & 4,163 & 3,054 \\
Q6. What is assured to the jury? & 18 & 7 & 2,552 & 4,699 \\
Q7. How is the right to association given? & 20 & 7 & 2,230 & 2,093 \\
Q8. How is legal assistance characterized for those with insufficient funds? & 12 & 8 & 5,689 & 1,523 \\
\bottomrule
\end{tabular}
\end{table}
 
\textbf{Figure} 9. Comparison of Max and Min Similarities for ML and
Flat Approaches

\includegraphics[width=6.31519in]{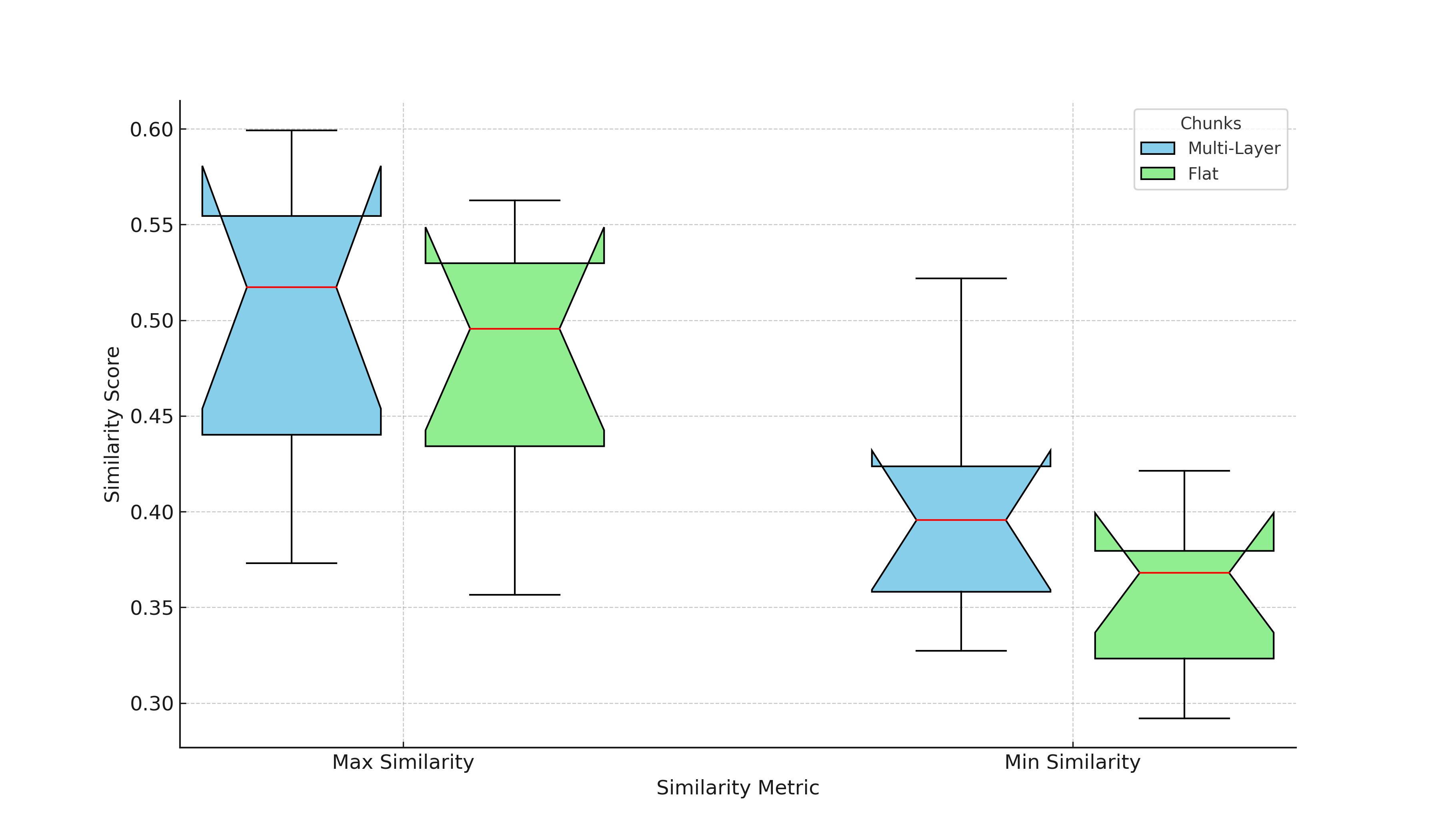}
\begin{multicols}{2}

Figure 9 presents a box plot with a comparative visualization of the
maximum and minimum similarity scores achieved by two distinct
techniques, namely Multi-Layer and Flat chunking approaches, across a
series of analytical questions. Each box plot distinctly represents the
interquartile range (IQR), medians, thus facilitating a clear overview
of the distribution patterns and variance within the data.

Upon close examination of the box plot comparing maximum and minimum
similarity scores for the Multi-Layer and Flat embedding approaches, it
is evident that the Multi-Layer approach tends to exhibit a narrower
range at the extremes. This observation is particularly significant as
it suggests that the chunks selected using the Multi-Layer embedding are
more semantically consistent and closely aligned with the user's query
embedding.

Figure 10 illustrates the classification results of chunks by relevance
(color-coded) and also displays cosine similarity between the question
and the chunks on the Y-axis. This visualization allows us to observe
how quickly the similarity decreases in the sequence of returned chunks.
Notably, only in two of the eight cases does the highest similarity
chunk coincide between the two chunking methods.

The Multi-Layer technique shows a higher proportion of essential chunks
(approximately 37.86\%) compared to the Flat Embedding method, which
accounts for only about 16.39\% of essential chunks. This suggests that
the Multi-Layer approach may be better at identifying critical content
in the texts, which could be advantageous for applications requiring
nuanced understanding of the document\textquotesingle s content.
Moreover, the Multi-Layer Embedding also has a lower proportion of
complementary chunks (around 3.88\%) compared to the Flat chunking
method (approximately 8.20\%). The Flat approach has a significantly
higher proportion of chunks classified as unnecessary (about 75.41\%)
compared to the Multi-Layer Embedding, which has about 58.25\% of chunks
in this category.

A particular challenge noted in both methodologies, but more effectively
managed in the multi-layered approach, is the semantic overload present
in Article 5 of the Brazilian Constitution. Questions 6, 7, and 8 (see
Appendix) highlight this issue, where the flat embedding struggles to
navigate the complex semantic layers within this article. The Article,
which covers a broad spectrum of fundamental rights, requires the
retrieval system to discern and prioritize among numerous embedded
concepts.

Figure 10. Questions and Chunk Relevance by Chunking Method

\includegraphics[width=3.28287in]{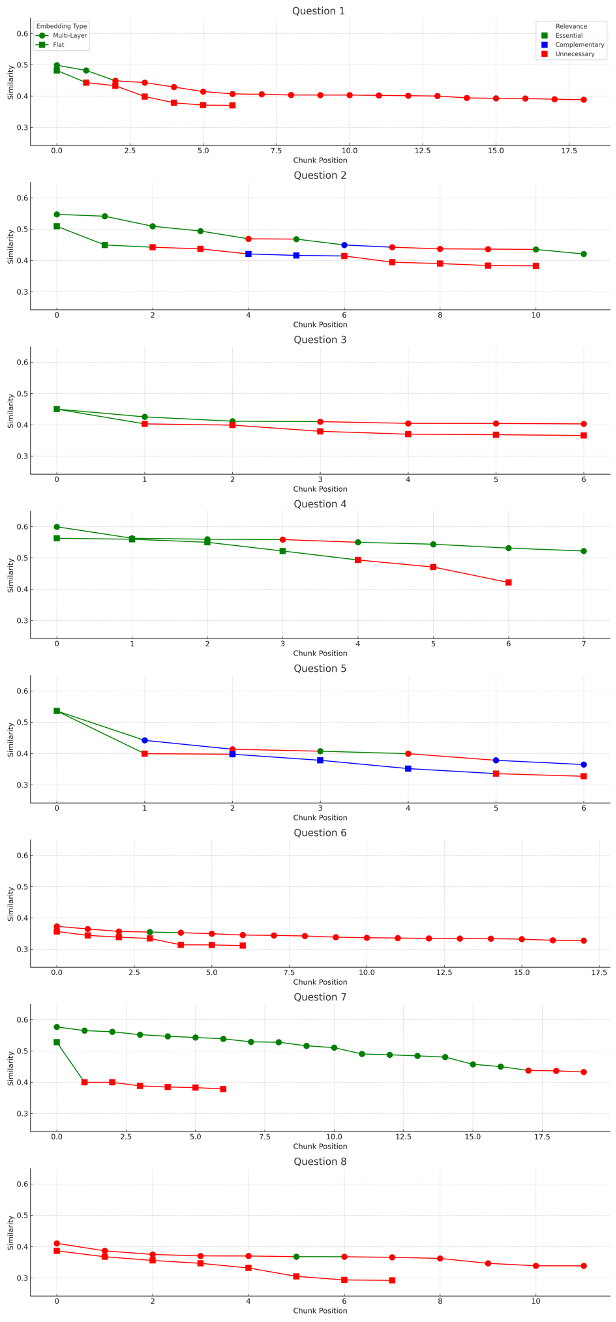}

\section{Conclusion}
\label{conclusion}

The proposed multi-layered embedding-based retrieval approach offers a
promising avenue for advancing legal information retrieval. By capturing
the inherent structure and diverse granularities of legal knowledge,
this method facilitates information access, ultimately empowering legal
professionals and researchers in their pursuit of legal knowledge. The
method's ability to represent legal texts at different levels of
granularity ensures that whether a query seeks an overview or a detailed
examination, the retrieval system can provide the LLM with the relevant
chunks as database source to create an answer that matches the user's
specific request.

The flexibility of question embeddings significantly enhances legal
searching by permitting queries in plain language or even in different
languages, narrowing the divide between legal practitioners and the
general public. A legal expert might pose a question in technical terms
such as `Where does power emanate from?' (`De onde emana o poder?') or
in Italian, \textquotesingle Da dove viene il potere?\textquotesingle.
Alternatively, a layperson asking `De onde vem a autoridade desse povo
que manda?' (Where do those in power get their authority ?) will receive
an equally accurate response. This capability illustrates the
system\textquotesingle s semantic depth---it prioritizes the collective
meaning over individual words.

The comparative evaluation between flat and multi-layered
embedding-based retrieval methodologies reveals significant differences
in the capacity to recognize and utilize essential text chunks. The flat
method often fails to identify critical segments that are crucial for
precisely answering queries. This limitation stems partly from its
inability to appreciate the full semantic relevance due to the absence
of broader contexts and additional layers of information.

In contrast, the multi-layered approach demonstrates a superior ability
in identifying and selecting relevant passages, owing to its text
analyses at various granularity levels. This not only enhances the
accuracy of the responses but also reflects a deeper comprehension of
the legal content, facilitating the retrieval of information that is
genuinely pertinent to the user\textquotesingle s inquiry. Moreover,
unlike the flat approach, the multi-layered methodology effectively
circumvents redundant or irrelevant chunks, which is critical for
maintaining the efficiency of legal information retrieval systems.

In conclusion, the adoption of a Retrieval Augmented Generation (RAG)
system with a multi-layered embedding approach can enhance the
efficiency and accuracy of legislative consultations. This technology
empowers consultants by enabling quicker access to pertinent legal
information and facilitating informed decision-making in the drafting
and legal analysis of legislation. By streamlining the retrieval and
utilization of legal texts, it improves the quality and precision of
legislative outputs, fostering a more effective and well-grounded
legislative process.

As we look toward future enhancements and expansions of this research,
several areas have been identified for further exploration and
development.

\begin{itemize}
\item
  Inter-Article Relationships: Investigate methods to represent
  relationships between articles, such as cross-references and
  dependencies. This could involve creating embeddings for pairs or
  groups of articles, capturing how they interact and inform each other.
\item
  Temporal Dimension: Consider incorporating a temporal dimension into
  the embeddings, reflecting the evolution of legal texts and amendments
  over time. This could involve creating embeddings for different
  versions of the legal text.
\item
  Vector Dimensions: During our proof-of-concept tests involving the
  Constitution we utilized 256 dimensions. Given that the adopted
  language model is capable of supporting up to 3072 dimensions, future
  investigations should assess whether increasing the number of
  dimensions can enhance performance even further. This exploration will
  help determine if the additional computational cost is justified by
  potentially improved accuracy and contextual sensitivity in the
  retrieval of complex legal documents, thus optimizing the trade-off
  between resource expenditure and retrieval efficacy.
\end{itemize}

The exploration of multi-layered embedding-based retrieval in legal
texts not only enhances the precision of information access but also
bridges the gap between complex legal terminologies and the general
public's understanding of them. As we continue to refine and expand upon
this methodology, the potential for scalable, linguistically diverse,
and contextually rich legal information systems seems increasingly
attainable. Future research will undoubtedly unlock further
capabilities, making legal knowledge more accessible and navigable for
professionals and laypersons alike.

\bibliographystyle{unsrt}
\bibliography{multilayer}

\section*{Acknowledgments}
The author is grateful to Ari Hershowitz, Lauro Cesar Araujo and João Paulo Almeida for their comments on the draft of this paper, which have contributed to its improvement. Additionally, the author thanks Luiza Rafaela Vasconcelos Chaffim for her support in formulating the questions and evaluating the relevance of the RAG responses.

\end{multicols}

\clearpage

\hypertarget{appendix-questions}{%
\section{Appendix -- Questions}\label{appendix-questions}}

\subsection*{Question 1. What are the foundations of the republic?}
 \begin{table}[H]
\centering
\small
\begin{tabular}{|p{7cm}|p{7cm}|}
\hline
\textbf{Multi-Layer -- GPT-4 Answer} & \textbf{Flat -- GPT-4 Answer} \\
\hline
The foundations of the Republic Federativa do Brasil, as outlined in the Constitution (CRFB), Art. 1º, are: sovereignty; citizenship; the dignity of the human person; the social values of labor and free enterprise; and political pluralism. & The foundations of the Republic Federativa do Brasil, as outlined in Art. 1º of the CRFB, include: \newline Sovereignty \newline Citizenship \newline The dignity of the human person \newline The social values of labor and free enterprise \newline Political pluralism \\
\hline
\end{tabular}
\end{table}

\begin{table}[H]
\centering
\begin{tabular}{|p{14cm}|}
\hline
\textbf{Evaluation of Answers and Chunks Selection} \\
\hline
Both the multi-layer chunking GPT-4 answer and the flat chunking GPT-4 answer effectively outline the foundations of the "Republic Federativa do Brasil," although the correct term in English would be "Federative Republic of Brazil." \newline

Several chunks beyond Art. 1º are marked as unnecessary (U), indicating that they do not directly contribute to understanding the foundational principles of the republic. These unnecessary chunks were correctly disregarded in the response generation phase, ensuring focus on the most pertinent information. \\
\hline
\end{tabular}
\end{table}

\begin{table}[H]
\centering
\small
\begin{minipage}[t]{0.48\textwidth} 
\centering
\caption{Q1 - Multi-Layer Chunking}
\begin{tabular}{|p{3.3cm}|c|c|c|}
\hline
\textbf{Label} & \textbf{Similarity} & \textbf{\# Tok} & \textbf{R} \\
\hline
Art. 1º, caput & 0.498642 & 82 & E \\
\hline
Art. 1º & 0.48199 & 134 & E \\
\hline
TÍTULO I & 0.448901 & 476 & U \\
\hline
Art. 89. & 0.443145 & 181 & U \\
\hline
Art. 29., caput, Inciso X & 0.428818 & 108 & U \\
\hline
Art. 170., caput,  Inciso I & 0.414425 & 66 & U \\
\hline
Art. 90., caput, \textit{Inciso II} & 0.406925 & 39 & U \\
\hline
Art. 29., caput, \textit{Inciso XIV} & 0.405706 & 117 & U \\
\hline
Art. 17., caput, \textit{Inciso I} & 0.403476 & 74 & U \\
\hline
TÍTULO IV, CAPÍTULO II, Seção V, Subseção I & 0.403246 & 316 & U \\
\hline
Art. 29., caput, \textit{Inciso XII} & 0.403231 & 106 & U \\
\hline
Art. 29., caput, \textit{Inciso I} & 0.402161 & 134 & U \\
\hline
Art. 34., caput, \textit{Inciso VII, Alínea a} & 0.401092 & 58 & U \\
\hline
Art. 17., caput, \textit{Inciso IV} & 0.400064 & 80 & U \\
\hline
Art. 29., caput, \textit{Inciso VIII} & 0.394153 & 127 & U \\
\hline
Art. 29., caput, \textit{Inciso XIII} & 0.392702 & 136 & U \\
\hline
Art. 29., caput, \textit{Inciso XI} & 0.392028 & 110 & U \\
\hline
Art. 170., caput, \textit{Inciso VII} & 0.389966 & 76 & U \\
\hline
Art. 17., caput, \textit{Inciso III} & 0.388281 & 82 & U \\
\hline
\end{tabular}
\end{minipage}%
\hfill
\begin{minipage}[t]{0.48\textwidth} 
\centering
\caption{Q1 - Flat Chunking}
\begin{tabular}{|p{3.3cm}|c|c|c|}
\hline
\textbf{Label} & \textbf{Similarity} & \textbf{\# Tok} & \textbf{R} \\
\hline
Art. 1º & 0.48199 & 134 & E \\
\hline
Art. 89. & 0.443145 & 181 & U \\
\hline
Art. 3º & 0.433032 & 109 & U \\
\hline
Art. 17. & 0.398566 & 125 & U \\
\hline
Art. 4º & 0.378643 & 180 & U \\
\hline
Art. 193. & 0.371011 & 94 & U \\
\hline
Art. 29. & 0.37029 & 2287 & U \\
\hline
\end{tabular}
\end{minipage}
\end{table}

Key: E – Essential; C – Complementary; U – Unnecessary 

\newpage

\subsection*{Question 2. Talk about the social function of property.}
\begin{table}[H]
\centering
\small
\begin{tabular}{|p{7cm}|p{7cm}|}
\hline
\textbf{Multi-Layer -- GPT-4 Answer} & \textbf{Flat -- GPT-4 Answer} \\
\hline
The social function of property is a principle deeply embedded in the Brazilian Constitution, reflecting a balance between individual property rights and the broader interests of society. This concept is articulated in several provisions of the Constitution, emphasizing the need for property, both rural and urban, to serve societal needs beyond mere ownership and control. \newline
In the context of rural property, Article 186 of the Constitution specifies that the social function of property is fulfilled when the property simultaneously meets certain criteria established by law. These criteria include the rational and adequate use of the property, the appropriate use of available natural resources and the preservation of the environment, compliance with regulations governing labor relations, and exploitation that benefits both the owners and the workers. This comprehensive approach underscores the expectation that rural properties contribute positively to the environment, economy, and social welfare. \newline
The Constitution also addresses the social function of property within the economic order. Article 170, clause III, explicitly states that the economic order, grounded on the valorization of human work and free initiative, aims to ensure a dignified existence for all according to the dictates of social justice, with the social function of property being one of its guiding principles. This highlights the integral role of property rights in achieving a just and equitable society, where economic activities and property ownership contribute to the collective well-being. \newline
Furthermore, the principle extends to urban property, as outlined in Article 182. The policy of urban development, executed by the municipal public power, aims to organize the full development of the city's social functions and ensure the well-being of its inhabitants. The social function of urban property is considered fulfilled when it meets the fundamental demands of city planning as expressed in the municipal master plan. This includes provisions for compulsory parceling or building on underused or unused urban land, progressive property taxes, and expropriation with compensation through public debt securities, emphasizing the importance of urban property serving the needs of urban development and habitation. \newline
Additionally, Article 5, clause XXIII, of the Constitution reinforces the concept by stating that property must fulfill its social function, embedding this principle within the broader framework of fundamental rights and guarantees. \newline
In summary, the social function of property in Brazil is a constitutional principle that mandates property, whether rural or urban, to contribute to the environmental, economic, and social goals of the nation. 

 & The social function of property is a fundamental principle enshrined in the Brazilian Federal Constitution (CRFB), reflecting the idea that property rights are not absolute and must serve broader societal goals beyond the interests of the individual property owner. This concept is articulated through various provisions in the Constitution, emphasizing the need for property to contribute to social well-being and justice. \newline
Article 186 of the CRFB specifically addresses the social function of rural property, stating that it is fulfilled when the property simultaneously meets certain criteria established by law. These criteria include the rational and adequate use of the property, the appropriate use of available natural resources and the preservation of the environment, compliance with regulations governing labor relations, and exploitation that benefits both the owners and the workers. \newline
Similarly, Article 182 outlines the social function of urban property within the context of urban development policy, executed by municipal governments. It mandates that urban property fulfills its social function by adhering to the fundamental demands of city planning as expressed in the master plan. This includes provisions for compulsory parceling or construction and progressive property taxes to encourage the proper utilization of urban land, aiming to ensure the well-being of city inhabitants. \newline
Moreover, Article 170 of the CRFB establishes the broader economic order, emphasizing the valorization of human work and free enterprise while ensuring a dignified existence for all under the principles of social justice. It highlights the social function of property as a key principle, alongside others such as national sovereignty, private property, free competition, consumer protection, and environmental defense. \newline
The concept of the social function of property is thus integral to the Brazilian legal and constitutional framework, balancing individual property rights with the necessity of promoting social justice, environmental sustainability, and economic development. It underscores the belief that property should not only serve the interests of its owners but also contribute to the collective well-being and development of society as a whole.
\\
\hline
\end{tabular}
\end{table}

\begin{table}[H]
\centering
\begin{tabular}{|p{14cm}|}
\hline
\textbf{Evaluation of Answers and Chunks Selection} \\
\hline
Both the multi-layer chunking and flat chunking GPT-4 answers provide a thorough understanding of the social function of property as established by the Brazilian Federal Constitution. However, it is noteworthy that the reference to Article 5, clause XXIII, which is essential for comprehensively understanding the social function of property, was included in the response and mentioned among the chunks only in the multi-layer approach.\newline

Several chunks beyond Art. 1º are marked as unnecessary (U), indicating that they do not directly contribute to understanding the foundational principles of the republic. These unnecessary chunks were correctly disregarded in the response generation phase, ensuring focus on the most pertinent information. \\
\hline
\end{tabular}
\end{table}

\begin{table}[H]
\centering
\small
\begin{minipage}[t]{0.48\textwidth} 
\centering
\caption{Q2 - Multi-Layer Chunking}
\begin{tabular}{|p{3.3cm}|c|c|c|}
\hline
\textbf{Label} & \textbf{Similarity} & \textbf{\# Tok} & \textbf{R} \\
\hline
CRFB, Art. 186., caput & 0,547111 & 105 & E \\
\hline
CRFB, Art. 170.,   caput, Inciso III & 0,541237   & 69     & E          \\
\hline
CRFB, Art. 186.                      & 0,509166   & 118    & E          \\
\hline
CRFB, Art. 170.,   caput, Inciso II  & 0,493896   & 68     & E          \\
\hline
CRFB, Art. 185., Parágrafo único.    & 0,46908    & 34     & U          \\
\hline
CRFB, Art. 5º,   caput, Inciso XXIII & 0,468134   & 91     & E          \\
\hline
CRFB, Art. 170.                      & 0,449425   & 243    & C          \\ 
\hline
CRFB, Art. 185.                      & 0,442181   & 103    & U          \\
\hline
CRFB, Art. 193.                      & 0,436986   & 94     & U          \\
\hline
CRFB, TÍTULO VIII, CAPÍTULO I        & 0,436138   & 105    & U          \\
\hline
CRFB, TÍTULO VII, CAPÍTULO III       & 0,435113   & 1204   & E          \\
\hline
CRFB, Art. 182.                      & 0,420642   & 352    & C          \\
\hline
\end{tabular}
\end{minipage}%
\hfill
\begin{minipage}[t]{0.48\textwidth} 
\centering
\caption{Q2 - Flat Chunking}
\begin{tabular}{|p{3.3cm}|c|c|c|}
\hline
\textbf{Label} & \textbf{Similarity} & \textbf{\# Tok} & \textbf{R} \\
\hline
CRFB, Art. 186. & 0,509166 & 118 & E \\
\hline
CRFB, Art. 170. & 0,449425 & 243 & E \\
\hline
CRFB, Art. 185. & 0,442181 & 103 & U \\
\hline
CRFB, Art. 193. & 0,436986 & 94 & U \\
\hline
CRFB, Art. 182. & 0,420642 & 352 & C \\
\hline
CRFB, Art. 184. & 0,416053 & 304 & C \\
\hline
CRFB, Art. 174. & 0,414197 & 266 & U \\
\hline
CRFB, Art. 243. & 0,394372 & 190 & U \\
\hline
CRFB, Art. 6º & 0,389956 & 144 & U \\
\hline
CRFB, Art. 203. & 0,383694 & 221 & U \\
\hline
CRFB, Art. 239. & 0,382548 & 590 & U \\
\hline
\end{tabular}
\end{minipage}
\end{table}

Key: E – Essential; C – Complementary; U – Unnecessary

\clearpage

\subsection*{Question 3. What are the attributes of the vote?}
\begin{table}[H]
\centering
\small
\begin{tabular}{|p{7cm}|p{7cm}|}
\hline
\textbf{Multi-Layer -- GPT-4 Answer} & \textbf{Flat -- GPT-4 Answer} \\
\hline
The attributes of the vote, as outlined in the provided context from the Constitution of the Federative Republic of Brazil (CRFB), are as follows: \newline
Universal: The vote is universal, meaning it is a right granted to all Brazilian citizens who meet certain criteria, without discrimination (CRFB, Art. 14). \newline
Direct: Voting is direct, with citizens casting their ballots directly for their candidates of choice without the intervention of electors or delegates (CRFB, Art. 14). \newline
Secret: The vote is secret, ensuring the privacy and confidentiality of voters' choices, protecting them from potential coercion or influence (CRFB, Art. 14; Art. 60, § 4º, Inciso II). \newline
Equal: Each vote has equal value, emphasizing the principle of equality among all voters (CRFB, Art. 14).
Periodic: The Constitution also implies the periodic nature of elections, ensuring regular intervals at which elections are held, though the specific term "periodic" is not directly mentioned in the provided chunks. The periodic nature is inherent in the democratic process and the stipulation against the abolition of direct, secret, universal, and periodic voting (CRFB, Art. 60, § 4º, Inciso II). \newline
These attributes are fundamental to the democratic process, ensuring that the sovereignty of the people is exercised in a fair, free, and equitable manner. 

 & The attributes of the vote, as outlined in CRFB, Art. 14, are as follows: the vote is exercised through universal suffrage and is direct and secret, with equal value for everyone.
\\
\hline
\end{tabular}
\end{table}

\begin{table}[H]
\centering
\begin{tabular}{|p{14cm}|}
\hline
\textbf{Evaluation of Answers and Chunks Selection} \\
\hline
Both the multi-layer and flat chunking GPT-4 answers detail the attributes of the vote as stated in the Constitution of the Federative Republic of Brazil, focusing on Article 14 and its stipulations about universal, direct, secret, equal, and periodic voting. The multi-layer answer is more comprehensive, incorporating references to Article 60, § 4º, Clause  II, which reinforces the constitutional safeguard against the abolition of these voting attributes. \\
\hline
\end{tabular}
\end{table}

\begin{table}[H]
\centering
\small
\begin{minipage}[t]{0.48\textwidth} 
\centering
\caption{Q3 - Multi-Layer Chunking}
\begin{tabular}{|p{3.3cm}|c|c|c|}
\hline
\textbf{Label} & \textbf{Similarity} & \textbf{\# Tok} & \textbf{R} \\
\hline
CRFB, Art. 14. & 0,450102 & 1024 & E \\
\hline
CRFB, Art. 60., §   4º, Inciso II & 0,425218 & 47 & E \\
\hline
CRFB, TÍTULO II, CAPÍTULO IV & 0,411463 & 1216 & E \\
\hline
CRFB, Art. 119., caput & 0,410047 & 110 & U \\
\hline
CRFB, Art. 93.,   caput, Inciso X & 0,404563 & 83 & U \\
\hline
CRFB, Art. 120., § 1º, Inciso I, Alínea b & 0,404346 & 65 & U \\
\hline
CRFB, Art. 73. & 0,402844 & 421 & U \\ 
\hline
\end{tabular}
\end{minipage}%
\hfill
\begin{minipage}[t]{0.48\textwidth} 
\centering
\caption{Q3 - Flat Chunking}
\begin{tabular}{|p{3.3cm}|c|c|c|}
\hline
\textbf{Label} & \textbf{Similarity} & \textbf{\# Tok} & \textbf{R} \\
\hline
CRFB, Art. 14. & 0,450102 & 1024 & E \\
\hline
CRFB, Art. 73. & 0,402844 & 421 & U \\
\hline
CRFB, Art. 119. & 0,398962 & 179 & U \\
\hline
CRFB, Art. 121. & 0,379178 & 360 & U \\
\hline
CRFB, Art. 120. & 0,369929 & 227 & U \\
\hline
CRFB, Art. 29. & 0,368285 & 2287 & U \\
\hline
CRFB, Art. 47. & 0,365846 & 51 & U \\ 
\hline
\end{tabular}
\end{minipage}
\end{table}

Key: E – Essential; C – Complementary; U – Unnecessary

\clearpage

\subsection*{Question 4. How is tax revenue distributed?}
\begin{table}[H]
\centering
\small
\begin{tabular}{|p{7cm}|p{7cm}|}
\hline
\textbf{Multi-Layer -- GPT-4 Answer} & \textbf{Flat -- GPT-4 Answer} \\
\hline
Tax revenue in Brazil is distributed among the Union, states, the Federal District, and municipalities according to specific rules set out in the Constitution of the Federal Republic of Brazil (CRFB). Here's a summary of the key points regarding the distribution:
Union to States and Municipalities: \newline
Art. 159: The Union delivers 50\% of the revenue collected from income taxes and industrialized products taxes in several ways, including: \newline
21.5\% to the Participation Fund for States and the Federal District.\newline
22.5\% to the Participation Fund for Municipalities.
3\% for financing programs in the North, Northeast, and Center-West regions.\newline
Additional 1\% to the Participation Fund for Municipalities in December, July, and September each year.\newline
10\% of the revenue from industrialized products taxes and certain other taxes goes to States and the Federal District based on the value of their exports.\newline
29\% of the revenue from the Contribution of Intervention in the Economic Domain goes to States and the Federal District, distributed as per law.\newline
States and the Federal District:\newline
Art. 157: States and the Federal District receive:
Revenue from income tax collected at the source on payments they make.\newline
20\% of the revenue from taxes the Union institutes under its exclusive powers.\newline
States to Municipalities:\newline
Art. 159, §3: States must pass on 25\% of the revenue they receive from the tax on industrialized products and certain other taxes to their municipalities, following specific criteria.\newline
Municipalities:\newline
Art. 158: Municipalities are entitled to:\newline
Revenue from income tax collected at the source on payments made by them.\newline
50\% of the revenue from the Union's tax on rural property within their territories.\newline
50\% of the revenue from the state tax on motor vehicles licensed in their territories.\newline
25\% of the revenue from the state tax on goods and services circulation and communication services.\newline
Art. 162: The Union, States, the Federal District, and Municipalities must disclose, by the last day of the month following the collection, the amounts of each tax collected, resources received, and the criteria for their distribution.\newline
Transparency and Reporting:\newline
This distribution mechanism ensures that tax revenue supports federal, state, and municipal levels of government, contributing to the development and maintenance of public services across the country.

 & Tax revenue distribution in Brazil is governed by a complex set of rules outlined in the Constitution of the Federative Republic of Brazil (CRFB). Here's a simplified overview based on the provided context:\newline
Federal Taxes: The Union, states, Federal District, and municipalities have the authority to institute taxes (CRFB, Art. 145). The distribution of revenue from federal taxes includes:\newline
A portion of the income tax and taxes on industrialized products is allocated to states and municipalities (CRFB, Art. 157, Art. 159, and Art. 158).\newline
States and the Federal District receive revenue from the Union's tax on income and earnings, specifically from amounts collected from their own institutions and from taxes the Union institutes under its exclusive powers (CRFB, Art. 157 and Art. 159).\newline
Municipalities receive a share of the income tax collected at the source on earnings paid by them, their autarchies, and foundations they establish and maintain. They also receive half of the revenue from the Union's tax on rural property located within their territories and a portion of the state tax on vehicle ownership (CRFB, Art. 158).\newline
Industrialized Products and Other Specific Taxes:
A share of the revenue from the tax on industrialized products and certain other federal taxes is distributed among states and municipalities, with specific percentages allocated to funds supporting regional development and municipal participation (CRFB, Art. 159).
\newline Transparency and Reporting:\newline
The Union, states, Federal District, and municipalities are required to disclose the amounts of each tax collected, the resources received, and the criteria for their distribution, ensuring transparency in the allocation of tax revenue (CRFB, Art. 162).\newline\newline
Social Security Financing:\newline
Social security is financed through contributions from employers, employees, and other sources, with revenues designated for health, pension, and social assistance programs. These contributions are distributed across federal, state, and municipal budgets (CRFB, Art. 195).\newline
Special Provisions:\newline
The Federal District is entitled to municipal taxes in Federal Territories, and if a Territory is not divided into municipalities, it also receives state taxes (CRFB, Art. 147).\newline
The tax system is designed to be fair, transparent, and efficient, with efforts to mitigate regressive effects and ensure that taxes and contributions are graded according to the economic capacity of the taxpayer (CRFB, Art. 145 and Art. 195).\newline
This framework ensures a multi-layered distribution of tax revenue, aiming to support federal, state, and municipal governments while promoting social welfare and regional development.

\\
\hline
\end{tabular}
\end{table}

\begin{table}[H]
\centering
\begin{tabular}{|p{14cm}|}
\hline
\textbf{Evaluation of Answers and Chunks Selection} \\
\hline
The multi-layer and flat GPT-4 answers comprehensively detail the distribution of tax revenue in Brazil as per the Constitution of the Federative Republic of Brazil (CRFB).\newline
The chunk selection in both the multi-layer and flat chunking approaches appropriately highlights the most relevant articles—Art. 157, Art. 159, Art. 158, and Art. 162—as essential for understanding tax revenue distribution. Note, however, that there is some repeated content in the chunks from the multi-layer approach, as more detailed and specific provisions precede more general ones of which they are part.
 \\
\hline
\end{tabular}
\end{table}

\begin{table}[H]
\centering
\small
\begin{minipage}[t]{0.48\textwidth} 
\centering
\caption{Q4 - Multi-Layer Chunking}
\begin{tabular}{|p{3.3cm}|c|c|c|}
\hline
\textbf{Label} & \textbf{Similarity} & \textbf{\# Tok} & \textbf{R} \\
\hline
CRFB, Art. 159., caput & 0,599326 & 487 & E \\
\hline
CRFB, Art. 157. & 0,562555 & 124 & E \\
\hline
CRFB, Art. 159. & 0,559442 & 850 & E \\
\hline
CRFB, Art. 212., § 1º & 0,558399 & 68 & U \\
\hline
CRFB, Art. 162. & 0,549713 & 128 & E \\
\hline
CRFB, Art. 158., caput & 0,543541 & 298 & E \\
\hline
CRFB, Art. 158., § 1º & 0,531194 & 170 & E \\
\hline
CRFB, Art. 158. & 0,521858 & 671 & E \\ 
\hline
\end{tabular}
\end{minipage}%
\hfill
\begin{minipage}[t]{0.48\textwidth} 
\centering
\caption{Q4 - Flat Chunking}
\begin{tabular}{|p{3.3cm}|c|c|c|}
\hline
\textbf{Label} & \textbf{Similarity} & \textbf{\# Tok} & \textbf{R} \\
\hline
CRFB, Art. 157. & 0,562555 & 124 & E \\
\hline
CRFB, Art. 159. & 0,559442 & 850 & E \\
\hline
CRFB, Art. 162. & 0,549713 & 128 & E \\
\hline
CRFB, Art. 158. & 0,521858 & 671 & E \\
\hline
CRFB, Art. 147. & 0,493165 & 62 & U \\
\hline
CRFB, Art. 145. & 0,470684 & 296 & U \\
\hline
CRFB, Art. 195. & 0,421283 & 1464 & U \\ 
\hline
\end{tabular}
\end{minipage}
\end{table}

Key: E – Essential; C – Complementary; U – Unnecessary

\clearpage

\subsection*{Question 5. What are the rights of children and teenagers?}
\begin{table}[H]
\centering
\small
\begin{tabular}{|p{7cm}|p{7cm}|}
\hline
\textbf{Multi-Layer -- GPT-4 Answer} & \textbf{Flat -- GPT-4 Answer} \\
\hline
The rights of children and teenagers in Brazil are comprehensively outlined in the Constitution of the Federative Republic of Brazil (CRFB). According to Art. 227 of the CRFB, children, teenagers, and young people are ensured, with absolute priority, the right to life, health, food, education, leisure, professionalization, culture, dignity, respect, freedom, and family and community living. Additionally, they are protected against all forms of negligence, discrimination, exploitation, violence, cruelty, and oppression. The State is mandated to promote comprehensive health care programs for children and teenagers, including the participation of non-governmental entities, focusing on maternal and child health services and specialized programs for individuals with physical, sensory, or mental disabilities.\newline
Furthermore, Art. 7, XXXIII of the CRFB prohibits night, hazardous, or unhealthy work for those under eighteen and any work for those under sixteen, except as apprentices starting from fourteen years of age. Art. 24, XV empowers the Union, states, and the Federal District to legislate concurrently on the protection of children and youth. The CRFB also emphasizes the family's role in providing a nurturing environment for children and teenagers, highlighting the State's duty to protect the family and ensure equal rights and qualifications for all children, regardless of their familial status (Art. 226 and Art. 227, §6).\newline
Education rights are further detailed in Art. 208, ensuring compulsory and free basic education for children and teenagers from four to seventeen years of age, along with specialized educational support for those with disabilities, and access to higher education based on individual capacity.\newline
In summary, the rights of children and teenagers in Brazil are deeply enshrined in the CRFB, covering a broad spectrum of protections and supports to ensure their well-being and development in society.\newline

 & The rights of children and teenagers in Brazil are comprehensively protected under the Constitution. According to Article 227 of the Constitution, children, adolescents, and young people are ensured, with absolute priority, the right to life, health, food, education, leisure, professionalization, culture, dignity, respect, freedom, and family and community living. Additionally, they are to be safeguarded against all forms of neglect, discrimination, exploitation, violence, cruelty, and oppression. The State is tasked with promoting programs for comprehensive health care for children and adolescents, including the participation of non-governmental entities, with specific policies such as allocating a percentage of public health resources to maternal and child health and creating programs for the prevention and specialized care for individuals with physical, sensory, or mental disabilities.\newline
Furthermore, Article 208 outlines the State's duty regarding education, ensuring the right to free and compulsory basic education from ages 4 to 17, progressively universalizing free secondary education, and providing specialized educational assistance to individuals with disabilities, preferably within the regular education system. It also mandates free early childhood education for children up to 5 years old and supports educational access at all levels according to each individual's capability.\newline
Article 228 establishes that individuals under 18 years of age are not criminally liable and are subject to special legislation.\newline
Moreover, Article 7 details the labor rights, including for young workers, such as protection against arbitrary dismissal, unemployment insurance, a national minimum wage sufficient for basic needs, and specific protections for young workers, including the prohibition of night, hazardous, or unhealthy work for those under 18 and any work for those under 16, except as apprentices from the age of 14.\newline
Article 6 recognizes education, health, food, work, housing, transportation, leisure, security, social welfare, protection of motherhood and infancy, and assistance to the destitute as social rights.\newline
Lastly, Articles 205 and 206 emphasize that education is a right for all and a duty of the State and the family, to be promoted with the society's collaboration, aiming at the full development of the individual, preparation for citizenship, and qualification for work. Education must be based on principles such as equal conditions for access and permanence in school, freedom of learning and teaching, and the guarantee of quality standards.\newline
These articles collectively ensure a broad spectrum of rights and protections for children and teenagers, emphasizing the importance of their well-being and development in Brazilian society.

\\
\hline
\end{tabular}
\end{table}

\begin{table}[H]
\centering
\begin{tabular}{|p{14cm}|}
\hline
\textbf{Evaluation of Answers and Chunks Selection} \\
\hline
Both the multi-layer and flat GPT-4 answers provide a comprehensive overview of the rights of children and teenagers in Brazil, as defined in the Constitution of the Federative Republic of Brazil (CRFB). Article 227, which is marked as essential in both approaches, is appropriately detailed in both answers, covering a wide array of rights including life, health, education, and protection against various forms of maltreatment. This article serves as the backbone for discussing children's and teenagers' rights and is crucial for a full understanding of the topic.\newline
Note that the chunk "CRFB, Art. 24., caput, Inciso XV," evaluated as unnecessary, was included in the response from the multi-layer approach. This inclusion, while not necessary to answer the question, does not necessarily indicate that it will be repeated in other generations of the same question using this approach.

 \\
\hline
\end{tabular}
\end{table}

\begin{table}[H]
\centering
\small
\begin{minipage}[t]{0.48\textwidth} 
\centering
\caption{Q5 - Multi-Layer Chunking}
\begin{tabular}{|p{3.3cm}|c|c|c|}
\hline
\textbf{Label} & \textbf{Similarity} & \textbf{\# Tok} & \textbf{R} \\
\hline
CRFB, Art. 227. & 0,535913 & 880 & E \\
\hline
CRFB, Art. 7º,   caput, Inciso XXXIII & 0,442036 & 92 & C \\
\hline
CRFB, Art. 24.,   caput, Inciso XV & 0,413621 & 40 & U \\
\hline
CRFB, TÍTULO VIII, CAPÍTULO VII & 0,407366 & 1420 & E \\
\hline
CRFB, Art. 208. & 0,399493 & 342 & C \\
\hline
CRFB, Art. 7º & 0,378321 & 1271 & C \\
\hline
CRFB, Art. 5º,   caput, Inciso XXVII & 0,364702 & 118 & U \\ 
\hline
\end{tabular}
\end{minipage}%
\hfill
\begin{minipage}[t]{0.48\textwidth} 
\centering
\caption{Q5 - Flat Chunking}
\begin{tabular}{|p{3.3cm}|c|c|c|}
\hline
\textbf{Label} & \textbf{Similarity} & \textbf{\# Tok} & \textbf{R} \\
\hline
CRFB, Art. 227. & 0,535913 & 880 & E \\
\hline
CRFB, Art. 208. & 0,399493 & 342 & C \\
\hline
CRFB, Art. 228. & 0,397702 & 33 & C \\
\hline
CRFB, Art. 7º & 0,378321 & 1271 & C \\
\hline
CRFB, Art. 6º & 0,351804 & 144 & C \\
\hline
CRFB, Art. 205. & 0,335678 & 66 & U \\
\hline
CRFB, Art. 206. & 0,327202 & 318 & U \\
\hline
\end{tabular}
\end{minipage}
\end{table}

Key: E – Essential; C – Complementary; U – Unnecessary

\clearpage

\subsection*{Question 6. What is assured to the jury?}
\begin{table}[H]
\centering
\small
\begin{tabular}{|p{7cm}|p{7cm}|}
\hline
\textbf{Multi-Layer -- GPT-4 Answer} & \textbf{Flat -- GPT-4 Answer} \\
\hline
The jury is assured the following according to Art. 5º, caput, Inciso XXXVIII of the CRFB:\newline
The fullness of defense (a plenitude de defesa);\newline
The secrecy of the votes (o sigilo das votações);\newline
The sovereignty of the verdicts (a soberania dos veredictos);\newline
The competence to judge willful crimes against life (a competência para o julgamento dos crimes dolosos contra a vida).
\newline

 & The Constitution of the Federative Republic of Brazil (CRFB) does not specifically mention or assure anything directly to "the jury" in the provided chunks. The excerpts focus on the responsibilities, immunities, and procedures related to various public officials and bodies, such as the President, members of Congress, public servants, and the organization of municipalities. If you are referring to legal protections or assurances for members of a jury in a judicial context, such information is not covered in the provided text excerpts.

\\
\hline
\end{tabular}
\end{table}

\begin{table}[H]
\centering
\begin{tabular}{|p{14cm}|}
\hline
\textbf{Evaluation of Answers and Chunks Selection} \\
\hline
The multi-layer GPT-4 answer accurately details the assurances provided to the jury as specified in Article 5, Inciso XXXVIII of the Constitution of the Federative Republic of Brazil (CRFB). It correctly lists the fundamental aspects assured to the jury, including the fullness of defense, secrecy of votes, sovereignty of the verdicts, and the competence to judge willful crimes against life. This response effectively utilizes the relevant chunk marked as essential, directly answering the question based on the constitutional provision.\newline
In contrast, the flat GPT-4 answer suggests that the provided chunks did not specifically mention or assure anything directly to "the jury." This indicates a potential misunderstanding or misalignment of the chunks in the flat chunking approach, as it failed to recognize or include the critical chunk from Article 5, Inciso XXXVIII, which is indeed relevant and contains essential assurances to the jury.\newline
This particular article of the Constitution, Article 5, holds significant semantic weight as it covers all fundamental rights—more than 70 in total. This breadth can lead to its non-selection in scenarios focusing only on one or a few fundamental rights, as is the case here.

 \\
\hline
\end{tabular}
\end{table}

\begin{table}[H]
\centering
\small
\begin{minipage}[t]{0.48\textwidth} 
\centering
\caption{Q6 - Multi-Layer Chunking}
\begin{tabular}{|p{3.3cm}|c|c|c|}
\hline
\textbf{Label} & \textbf{Similarity} & \textbf{\# Tok} & \textbf{R} \\
\hline
CRFB, Art. 41., §   1º, Inciso I & 0,373043 & 37 & U \\
\hline
CRFB, Art. 41., §   1º, Inciso II & 0,36449 & 42 & U \\
\hline
CRFB, Art. 86. & 0,356619 & 274 & U \\
\hline
CRFB, Art. 5º,   caput, Inciso XXXVIII & 0,354712 & 156 & E \\
\hline
CRFB, Art. 14., § 11. & 0,352659 & 44 & U \\
\hline
CRFB, Art. 52., Parágrafo único. & 0,349301 & 92 & U \\
\hline
CRFB, Art. 128.,   § 5º, Inciso I & 0,345123 & 235 & U \\
\hline
CRFB, Art. 53. & 0,344011 & 490 & U \\
\hline
CRFB, Art. 121., § 1º & 0,342131 & 58 & U \\
\hline
CRFB, Art. 121. & 0,338429 & 360 & U \\
\hline
CRFB, Art. 247., Parágrafo único. & 0,336543 & 51 & U \\
\hline
CRFB, Art. 167-G., caput & 0,335619 & 47 & U \\
\hline
CRFB, Art. 41. & 0,334286 & 291 & U \\
\hline
CRFB, Art. 129.,   caput, Inciso I & 0,334219 & 41 & U \\
\hline
CRFB, Art. 29.,   caput, Inciso VIII & 0,33355 & 127 & U \\
\hline
CRFB, Art. 73., §   1º, Inciso II & 0,331975 & 54 & U \\
\hline
CRFB, Art. 95.,   caput, Inciso II & 0,328378 & 45 & U \\
\hline
CRFB, Art. 29.,   caput, Inciso X & 0,327375 & 108 & U \\ 
\hline
\end{tabular}
\end{minipage}%
\hfill
\begin{minipage}[t]{0.48\textwidth} 
\centering
\caption{Q6 - Flat Chunking}
\begin{tabular}{|p{3.3cm}|c|c|c|}
\hline
\textbf{Label} & \textbf{Similarity} & \textbf{\# Tok} & \textbf{R} \\
\hline
CRFB, Art. 86. & 0,356619 & 274 & U \\
\hline
CRFB, Art. 53. & 0,344011 & 490 & U \\
\hline
CRFB, Art. 121. & 0,338429 & 360 & U \\
\hline
CRFB, Art. 41. & 0,334286 & 291 & U \\
\hline
CRFB, Art. 52. & 0,313872 & 854 & U \\
\hline
CRFB, Art. 247. & 0,313551 & 143 & U \\
\hline
CRFB, Art. 29. & 0,311368 & 2287 & U \\ 
\hline
\end{tabular}
\end{minipage}
\end{table}

Key: E – Essential; C – Complementary; U – Unnecessary

\clearpage

\subsection*{Question 7. How is the right to association given?}
\begin{table}[H]
\centering
\small
\begin{tabular}{|p{7cm}|p{7cm}|}
\hline
\textbf{Multi-Layer -- GPT-4 Answer} & \textbf{Flat -- GPT-4 Answer} \\
\hline
The right to association is provided under the Brazilian Constitution in several aspects, ensuring a broad protection of this fundamental right. According to Art. 5, caput, Inc. XVII, there is full freedom of association for lawful purposes, with the prohibition of associations of a paramilitary nature. Additionally, Inc. XX of the same article guarantees that no one can be compelled to associate or to remain associated, emphasizing the voluntary nature of associations.\newline
Furthermore, the Constitution extends specific provisions regarding the formation and operation of associations and unions. For instance, Art. 8, caput, outlines the freedom of professional or union association, detailing rights such as the prohibition of requiring state authorization for the foundation of a union, subject only to registration (Inc. I); the defense of collective or individual rights of the category by the union (Inc. III); and the protection against arbitrary dismissal of an employee unionized from the moment of registration of candidacy for union leadership positions until one year after the end of their mandate, unless in cases of gross misconduct (Inc. VIII).\newline
Moreover, Art. 5, caput, Inc. XVIII, states that the creation of associations and cooperatives does not depend on authorization, reinforcing the autonomy and freedom from state interference in their functioning. This framework is designed to foster a civil society where individuals can freely organize around shared interests, contributing to democratic participation and the defense of collective and individual rights.

 & The right to association, particularly in the context of professional or union associations, is provided under the Constitution of the Federative Republic of Brazil (CRFB) as a free right, with specific regulations outlined mainly in Article 8. According to this article, individuals are free to form unions without requiring authorization from the State, although registration with the competent body is necessary. The public power is prohibited from interfering in or intervening with union organization. Additionally, the creation of more than one union representing the same professional or economic category within the same territorial base, which cannot be smaller than the area of a municipality, is forbidden. Unions are tasked with defending the collective or individual rights and interests of their category, including in judicial or administrative matters. The article also establishes that union contributions, decided by general assembly, may be deducted from members' salaries to fund the system of syndical representation, independent of legally mandated contributions. Importantly, no one is obligated to join or remain a member of a union, but unions must participate in collective labor negotiations. The protection extends to elected union representatives, who cannot be dismissed from the time of their candidacy until one year after their term ends, unless they commit a severe fault as defined by law. These provisions also apply to rural unions and fishermen's colonies, with specific conditions set by law.

\\
\hline
\end{tabular}
\end{table}

\begin{table}[H]
\centering
\begin{tabular}{|p{14cm}|}
\hline
\textbf{Evaluation of Answers and Chunks Selection} \\
\hline
Both the multi-layer and flat GPT-4 answers accurately address the right to association as outlined in the Constitution of the Federative Republic of Brazil (CRFB), highlighting different aspects and provisions that frame this fundamental right. \newline
The multi-layer answer provides a detailed and nuanced explanation by discussing various relevant articles such as Article 5, which encompasses the general principles of association, and Article 8, which focuses specifically on professional and union associations. This answer effectively incorporates a wider array of articles, demonstrating a comprehensive understanding of the topic.\newline
The flat answer focuses more specifically on professional and union associations as detailed in Article 8. It highlights the freedoms and regulations pertaining to union formation and operation, providing a clear overview of the specific constitutional guarantees that protect the right to association in a professional context.\newline
Similar to Question 6, the semantic overload of Article 5 resulted in it not being selected by the flat chunking approach. In contrast, the multi-layer approach selected eight clauses from Article 5, enabling a more comprehensive response.

 \\
\hline
\end{tabular}
\end{table}

\begin{table}[H]
\centering
\small
\begin{minipage}[t]{0.48\textwidth} 
\centering
\caption{Q7 - Multi-Layer Chunking}
\begin{tabular}{|p{3.3cm}|c|c|c|}
\hline
\textbf{Label} & \textbf{Similarity} & \textbf{\# Tok} & \textbf{R} \\
\hline
CRFB, Art. 8º,   caput, Inciso V & 0,576668 & 48 & E \\
\hline
CRFB, Art. 8º,   caput, Inciso VII & 0,564867 & 51 & E \\
\hline
CRFB, Art. 8º,   caput, Inciso IV & 0,561134 & 83 & E \\
\hline
CRFB, Art. 8º,   caput, Inciso II & 0,552001 & 91 & E \\
\hline
CRFB, Art. 8º,   caput, Inciso VI & 0,54675 & 48 & E \\
\hline
CRFB, Art. 8º,   caput, Inciso I & 0,543053 & 77 & E \\
\hline
CRFB, Art. 8º,   caput, Inciso VIII & 0,538775 & 91 & E \\
\hline
CRFB, Art. 8º,   caput, Inciso III & 0,529025 & 60 & E \\
\hline
CRFB, Art. 8º & 0,527786 & 429 & E \\
\hline
CRFB, Art. 5º,   caput, Inciso XVII & 0,516435 & 106 & E \\
\hline
CRFB, Art. 5º,   caput, Inciso XXI & 0,510463 & 110 & E \\
\hline
CRFB, Art. 5º,   caput, Inciso LXX, Alínea b & 0,490327 & 137 & E \\
\hline
CRFB, Art. 5º,   caput, Inciso LXX & 0,487655 & 150 & E \\
\hline
CRFB, Art. 5º,   caput, Inciso XIX & 0,484264 & 124 & E \\
\hline
CRFB, Art. 5º,   caput, Inciso XX & 0,480439 & 97 & E \\
\hline
CRFB, Art. 5º,   caput, Inciso XVIII & 0,457267 & 116 & E \\
\hline
CRFB, Art. 5º,   caput, Inciso XVI & 0,450047 & 139 & E \\
\hline
CRFB, Art. 7º,   caput, Inciso I & 0,437415 & 83 & U \\
\hline
CRFB, Art. 10., caput & 0,436476 & 51 & U \\
\hline
CRFB, Art. 5º,   caput, Inciso XXVIII, Alínea b & 0,432799 & 139 & U \\
\hline
\end{tabular}
\end{minipage}%
\hfill
\begin{minipage}[t]{0.48\textwidth} 
\centering
\caption{Q7 - Flat Chunking}
\begin{tabular}{|p{3.3cm}|c|c|c|}
\hline
\textbf{Label} & \textbf{Similarity} & \textbf{\# Tok} & \textbf{R} \\
\hline
CRFB, Art. 8º & 0,527786 & 429 & E \\
\hline
CRFB, Art. 11. & 0,399731 & 53 & U \\
\hline
CRFB, Art. 10. & 0,39969 & 56 & U \\
\hline
CRFB, Art. 240. & 0,388288 & 66 & U \\
\hline
CRFB, Art. 217. & 0,384761 & 248 & U \\
\hline
CRFB, Art. 17. & 0,382792 & 934 & U \\
\hline
CRFB, Art. 226. & 0,37849 & 307 & U \\ 
\hline
\end{tabular}
\end{minipage}
\end{table}

Key: E – Essential; C – Complementary; U – Unnecessary

\clearpage

\subsection*{Question 8. How is legal assistance characterized for those with insufficient funds?}
\begin{table}[H]
\centering
\small
\begin{tabular}{|p{7cm}|p{7cm}|}
\hline
\textbf{Multi-Layer -- GPT-4 Answer} & \textbf{Flat -- GPT-4 Answer} \\
\hline
Legal assistance for those with insufficient funds is characterized as integral and free, as stated in Art. 5, caput, Inciso LXXIV of the CRFB. This provision ensures that the state will provide comprehensive legal aid at no cost to individuals who can prove a lack of financial resources. Furthermore, the Defensoria Pública (Public Defender's Office) is identified as a permanent institution essential to the judicial function of the state, responsible for providing legal guidance, promoting human rights, and defending, in all degrees, judicial and extrajudicial, the individual and collective rights of those in need, as outlined in Art. 134, caput, and further detailed in its subsequent sections. This service is fundamentally aimed at assisting the needy in a complete and free manner, as reinforced by the same article.

 & Legal assistance for those with insufficient funds is characterized by the provision of integral and free services, as outlined in Art. 134 of the CRFB. This article establishes the Defensoria Pública (Public Defender's Office) as a permanent institution essential to the judicial function of the State. It is tasked with providing legal guidance, promoting human rights, and defending, in all degrees, both judicially and extrajudicially, the individual and collective rights of those in need. This service is provided in accordance with the principles of democracy, ensuring access to justice for all, especially for those who demonstrate a lack of resources.

\\
\hline
\end{tabular}
\end{table}

\begin{table}[H]
\centering
\begin{tabular}{|p{14cm}|}
\hline
\textbf{Evaluation of Answers and Chunks Selection} \\
\hline
Both the multi-layer and flat GPT-4 answers comprehensively address the characterization of legal assistance for those with insufficient funds as outlined in the Constitution of the Federative Republic of Brazil (CRFB). Each response effectively highlights the essential role of the Defensoria Pública (Public Defender's Office) in providing free and integral legal services to those who lack financial resources. \newline
The multi-layer answer provides a detailed discussion by referring to both Article 5, Inciso LXXIV, which establishes the right to free legal assistance, and Article 134, which details the responsibilities and functions of the Public Defender's Office. This answer leverages the relevant legal frameworks to give a complete picture of how legal assistance is administered.\newline
The flat answer, while concise, effectively summarizes the provisions of Article 134, focusing on the operational aspects of the Public Defender's Office and its role in promoting human rights and providing legal assistance. However, it omits direct reference to Article 5, Inciso LXXIV, which explicitly states the right to free legal assistance, somewhat narrowing the scope of the response compared to the multi-layer answer.

 \\
\hline
\end{tabular}
\end{table}

\begin{table}[H]
\centering
\small
\begin{minipage}[t]{0.48\textwidth} 
\centering
\caption{Q8 - Multi-Layer Chunking}
\begin{tabular}{|p{3.3cm}|c|c|c|}
\hline
\textbf{Label} & \textbf{Similarity} & \textbf{\# Tok} & \textbf{R} \\
\hline
CRFB, Art. 24.,   caput, Inciso XIII & 0,41044 & 40 & U \\
\hline
CRFB, Art. 213. & 0,386464 & 281 & U \\
\hline
CRFB, TÍTULO IV, CAPÍTULO IV, Seção III & 0,374871 & 52 & U \\
\hline
CRFB, Art. 134., caput & 0,370232 & 114 & U \\
\hline
CRFB, Art. 227.,   § 3º, Inciso VI & 0,369935 & 85 & U \\
\hline
CRFB, Art. 5º,   caput, Inciso LXXIV & 0,367759 & 105 & E \\
\hline
CRFB, Art. 134. & 0,367547 & 401 & U \\
\hline
CRFB, TÍTULO IV, CAPÍTULO IV, Seção IV & 0,365672 & 453 & U \\
\hline
CRFB, Art. 5º,   caput, Inciso LXXVI & 0,362216 & 117 & U \\
\hline
CRFB, Art. 245. & 0,346508 & 64 & U \\
\hline
CRFB, Art. 5º,   caput, Inciso LXXIII & 0,338711 & 174 & U \\
\hline
CRFB, TÍTULO IV, CAPÍTULO IV & 0,33843 & 3803 & U \\
\hline
\end{tabular}
\end{minipage}%
\hfill
\begin{minipage}[t]{0.48\textwidth} 
\centering
\caption{Q8 - Flat Chunking}
\begin{tabular}{|p{3.3cm}|c|c|c|}
\hline
\textbf{Label} & \textbf{Similarity} & \textbf{\# Tok} & \textbf{R} \\
\hline
CRFB, Art. 213. & 0,386464 & 281 & U \\
\hline
CRFB, Art. 134. & 0,367547 & 401 & U \\
\hline
CRFB, Art. 133. & 0,35573 & 46 & U \\
\hline
CRFB, Art. 245. & 0,346508 & 64 & U \\
\hline
CRFB, Art. 199. & 0,331805 & 238 & U \\
\hline
CRFB, Art. 131. & 0,304878 & 240 & U \\
\hline
CRFB, Art. 167-C. & 0,293365 & 176 & U \\
\hline
CRFB, Art. 197. & 0,292082 & 77 & U \\ 
\hline
\end{tabular}
\end{minipage}
\end{table}

Key: E – Essential; C – Complementary; U – Unnecessary

\end{document}